\documentclass[journal]{IEEEtran}

\usepackage[pdftex]{graphicx}
\DeclareGraphicsExtensions{.pdf, .png, .jpg, .bmp}
\usepackage{subfigure}

\usepackage{blindtext}
\usepackage{dblfloatfix} 
\usepackage{multicol}
\usepackage{multirow}

\usepackage[cmex10]{amsmath}
\usepackage{amssymb}
\usepackage{array}

\usepackage[ruled,,vlined,lined,linesnumbered]{algorithm2e}

\newtheorem{definition}{Definition}

\begin{document}

\title{Edge Detection based on Kernel \\Density Estimation}

\author{Osvaldo Pereira, Esley Torre, Yasel Garc\'es and Roberto Rodr\'iguez}
        

%
%

\maketitle

\begin{abstract}
Edges of an image are considered a  crucial type of information. These can be extracted by applying edge detectors with different methodology. Edge detection is a vital step in computer vision tasks, because it is an essential issue for pattern recognition and visual interpretation. In this paper, we propose a new method for edge detection in images, based on the estimation by kernel of the probability density function. In our algorithm, pixels in the image with minimum value of density function are labeled as edges. The boundary between two homogeneous regions is defined in two domains: the spatial/lattice domain and the range/color domain. Extensive experimental evaluations proved that our edge detection method is significantly a competitive algorithm.
\end{abstract}


\section{Introduction}
In analysis systems and image processing is essential to distinguish among objects of interest and the rest of the image. The used techniques in order to determine the objects of interest are known as image segmentation. One of the most common is the segmentation by edge detection.

An edge can be defined as a significant change in the value of the
pixel intensity in a region of the image \cite{FreiChen1977}. The main purpose of the edge detection is to simplify the image data in order to minimize the amount of information to be processed  \cite{Canny1986}.

Generally, an edge is defined as the boundary pixels that connect two separate regions \cite{FreiChen1977,GonzalezWoods2001,Pratt1991}. The detection operation starts with the examination of the local discontinuity at each pixel element at the image. Amplitude, orientation and location of a particular subarea are the main characteristics of possible edges \cite{FreiChen1977}. Based on these characteristics, the edge detector must to decide whether each of the examined pixels is an edge or not.

Classical edge detection methods labeled a pixel as edge according to discontinuities in gray levels, colors or textures. The Roberts \cite{Roberts1965}, Sobel  \cite{Sobel1973}, and Prewitt \cite{Prewitt1970} operators detect edges by convolving a grayscale image with local derivative filters. Marr and Hildreth \cite{MarrHildreth1980} used zero crossings of the Laplacian of Gaussian operator. The Canny detector \cite{Canny1986} also models edges as sharp discontinuities in the brightness channel, adding non-maximum suppression and hysteresis thresholding steps.

There are many other techniques in the literature used for edge detection like \cite{Lakshmi2010} and \cite{Papari2011}; some of them are based on histograms, error minimization, maximizing an object function, fuzzy logic, wavelet approach, morphology, genetic algorithms, neural network and among others. 

A method for image segmentation by active contours has been proposed in \cite{Michailovich2007}, it based on a variational analysis, in which the active contour is driven by the forces stemming from minimization of a cost functional. The segmentation method  proposed by \cite{Michailovich2007} is based on a distance among probability densities. In particular, the active contours have been evolved to maximize the Bhattacharyya distance among nonparametric (kernel-based) estimates of the probability densities of segmentation classes \cite{Bhattacharyya67}. 

One algorithm that makes use of nonparametric density estimation is Mean shift (MSH) \cite{Fukunaga75}, \cite{Shen75}, \cite{Cheng95}. In essence, MSH is an iterative mode detection algorithm in the density distribution space  \cite{Shen75}, \cite{Comaniciu2002}, \cite{Comaniciu2000}. 

In \cite{AliHassan2012}, the authors proposed an approach for Mean Shift algorithm and contour image. The technique was applied to CT Angiography images. Another approach proposed an algorithm that constructs a kernel function histogram combining intensity and then uses the Mean Shift algorithm with this kernel function in order to detect automatically  edges in the gray level image \cite{Zhengzhou13}.


We find  in \cite{Economou2001} an adaptive algorithm to accomplish edge detection in multidimensional and color images. This is a statistical approach based on local work and nonparametric kernel density estimation. The location of the edge discontinuity coincides with the minimum of the image density function  and it is determined by an appropriate resampling of the locally  defined space of probability. 

In this paper, we propose a new method for edge detection based on the kernel estimation of the probability density function. As in \cite{Economou2001},  in our approach the pixels in the image with the minimum value of the density function are labeled as edges.

The main difference between \cite{Economou2001} and our method is that: in \cite{Economou2001}, the boundary between two homogeneous regions is defined in terms only of gray level domain, while that in our algorithm we uses the range-spatial domain (gray levels and  pixel position).

The remainder of the paper is organized as follows. In Section II, the theorical aspetcs concerning the kernel density estimation are exposed. Section III, describes in details our edge detections algorithm. The experimental results, comparisons and discussion are presented in Section IV. In section V, the most important conclusions are given.

\section{Theorical Aspects}
\subsection{Kernel density estimation}
One of the most popular nonparametric density estimators is estimation by density kernels. Mathematically speaking, the general multivariate kernel density estimate at the point $x$, is defined by:
\begin{equation}\label{eq:eq_4}
\hat{f}(x) =\frac{1}{nh^d}\sum\limits_{i=1}^{n}K\left(\frac{x-x_{i}}{h}\right)
\end{equation}
where $n$ data points $x_{i} , i= 1, 2, 3, \ldots, n$, represent a population with some unknown density function $f(x)$ \cite{Fukunaga75}, \cite{Cheng95}, \cite{Comaniciu2000}. 

For image segmentation, the feature space is composed of two independent domains: the $spatial/lattice$ domain and the $range/color$ domain. 
Due to the different natures of the domains, the kernel is usually broken into the product of two different radially symmetric kernels: 
\begin{equation}\label{eq:eq_6}
\hat{f}(x) = \frac{c}{n(h_{s})^{p}(h_{r})^{q}}\sum\limits_{i=1}^{n}k_{s}
\displaystyle{
\left(\begin{Vmatrix}
\frac{x - x_{i}}{h_{s}}
\end{Vmatrix}^2\right)
k_{r}
\left(\begin{Vmatrix}
\frac{x - x_{i}}{h_{r}}
\end{Vmatrix}^2\right)
}
\end{equation}
where $x$ is a pixel, $k_{s}$ and $k_{r}$ are the profiles used in the two respective domains, $h_{s}$ and $h_{r}$ are employed bandwidths in spatial-range domains and $c$ is the normalization constant.

As was shown in (\ref{eq:eq_6}), it there are two main parameters that have to be defined by the user: the spatial bandwidth $h_{s}$ and the range bandwidth $h_{r}$.
%
%

\subsection{Validation of Edge Detection}
At the present time, an unique segmentation method that achieves good results for any image type does not exist. For this reason, it is necessary to quantify the efficiency of an edge detection method comparing the obtained results with a real
model \cite{Barra2007}, \cite{Bouix2007},\cite{Crum}. However, to find a measure that carries out a correct evaluation of the obtained borders is a complex problem \cite{Christian2010}. 

Many techniques have been proposed for evaluation of edge detection algorithms. One of them is the Rand Index \cite{Rand1971}, which was introduced for a general clustering evaluation. The Rank Index operates by comparing the compatibility of assignments between pairs of elements in the clusters. 

Acoording to \cite{Rand1971} and \cite{Lawrence1985}, the rank index is:

%


\begin{equation}
RI = \frac{a+b}{a+b+c+d} = \frac{a+b}{\binom{n}{2}}
\end{equation}
where $a + b$ is the number of agreements between $X$ and $Y$ and $c + d$ is the number of disagreements between X and Y.

Variants of the Rand Index have been proposed to deal with the case of multiple ground-truth segmentations \cite{Unnikrishnan2007}, \cite{Yang2008}. Given a set of ground-truth segmentations ${G_k}$, the Probabilistic Rand Index (PRI) is defined as:
\begin{equation}
\label{eq:eq_11}
PRI(S,\lbrace G_k\rbrace) = \frac{1}{T}\sum\limits_{i<j}[c_{ij}p_{i,j} + (1-c_{ij})(1-p_{ij})]
\end{equation}
where $c_{ij}$ is the event that pixels $i$ and $j$ have the same label, $p_{ij}$ is the probability of the event and the $T$ letter is the total number of pixel pairs. 

The PRI has the drawback of suffering of a small dynamic range \cite{Unnikrishnan2007}, \cite{Yang2008}. In \cite{Unnikrishnan2007},  this drawback is resolved with normalization in order to produce the Normalized Probabilistic Rand Index (NPRI). The NPRI uses a typical normalization scheme: if the baseline index value is the expected value of the index of any given segmentation of a particular image, then:
\begin{equation}
NPRI = \frac{PRI - Expected Index}{Maximum Index - Expected Index}
\end{equation}

Recently another metric of similarity among images was proposed in \cite{Garces2013}. The Natural Entropy Distance (NED) was introduced for the purpose of comparing two images. NED is an index of similarity among images that use $\mathbb{Z}_n$ rings and the entropy function, this defined as:
\begin{definition}
Let $A$ and $B$ be, two images; then the natural entropy distance is defined by
\begin{equation}
\hat{\nu}(A, B) = E(A + (-B))
\end{equation}
where $-(B)$ is the additive inverse of $B$ and this is calculated by using the inverse of each pixel of $B$ in $\mathbb{Z}_n$.
\end{definition}

This index was applied as new stopping criterion to the Mean Shift Iterative Algorithm ($MSHi$) with the goal of reaching a better segmentation. The properties of this index were demostrated in \cite{Garces2013}. Some of them are the non-negativity, symmetry, invariance under affine transformations, such as: translation, reflection and rotation. Also, this fulfills with the axiom of identity of indiscernibles.

\section{Edge detection using a Probability Density Function}

As it was pointed out in \cite{Gonzalez2002}, the most common way of an edges detector is by convolving the image with a mask. The response of the mask at any point of the image is given by:
\begin{equation}
R = z_{1}\cdot f_{1}+z_{2}\cdot f_{2}+ \ldots + z_{9}\cdot f_{9} = \sum\limits_{i = 1}^{9} z_{i}\cdot f_{i} 
\end{equation} 
where, $f_{i}$ is the gray level of the pixel in the image directly below coefficient $z_{i}$ in the used mask. The value of $R$ is assigned to the central pixel of the mask in the output image.

If the response of the mask into the central position satisfies expression (\ref{eq:umbral}), then we can say that we have found an edge point.
\begin{equation}\label{eq:umbral}
\vert R \vert < u
\end{equation}
where $u$ is a nonnegative threshold.

Our approach for edge detection, which it is named ``Edge Detection by Density" (EDD)(see Algorithm \ref{DensityAlgorithm}), it is based on the estimation of density function in the central pixel of the mask, where the results of the convolution are stored into a new image. The obtained density image has values of pixels that belong to the interval $\left[0,1\right]$.

After this process, it is necessary to use a threshold, which is obtained according to the following steps (see Figure 1):
\begin{enumerate}
\item Calculate the histogram of the density image.
\item To find the threshold ($u$), around the associated density value of the biggest value of frequency in the histogram.
\item Label as edges all pixels with the lower density values than threshold $u$.
\end{enumerate}

Computationally speaking, a linear implementation of our method over CPU has an algorithmic complexity of $\Theta(hn)$, where $h$ is a kernel bandwidth and $n$ is the number of pixels in the image. A parallel version of the algorithm over GPU has $\Theta(1)$ of algorithmic complexity.

From Figure \ref{fig:AlgIterations}(b) to Figure \ref{fig:AlgIterations}(k), some algorithm iterations over the quadrant of Figure \ref{fig:AlgIterations}(a) are shown. Note that an implicit edge there is, which this separates two regions  of different labels. 
\begin{algorithm}[h]
\caption{Edge Detection by Density.}
\label{DensityAlgorithm}
\KwData{\\
	$I$: input image\; 
	$m$: image width; $n$: image height\;
	$h_{s}$: bandwidths in $spatial/lattice$ domain\;
	$h_{r}$: bandwidths in $range/color$ domain\;	
}
Initialize:\\
$R=0$\;
\For{$i=1,2,\ldots, m$}{
	\For{$j=1, 2,\ldots, n$}{		
		$R(i, j) = \hat{f}(p(i,j), h_{s}, h_{r})$\; 
		where $p$ is the pixel in the Image $I$ at position $i, j$ 
		and $\hat{f}$ is the kernel density estimation funtion (\ref{eq:eq_6}).  				
	}
}
$u = threshold(R)$
$R = \vert R \vert < u$ \;
\KwResult{$R$ is the edges image.}
\end{algorithm}

Figure \ref{fig:AlgIterations}(l) shows the obtained edges at the end of the algorithm. For this example two kernels were combined, the uniform kernel 
\begin{equation}
K_s(x)= \left\{
\begin{array}{lcl}
	1 & \mbox{ if } & \| x\| \leq 0 \\	
	0 & \mbox{ if } & \| x\| \geq 0
\end{array} 
\right.
\end{equation}
and Gaussian kernel   
\begin{equation}
K_r(x) = e^{-\| x\|^2}
\end{equation}

Many researches have been carried out for edge detection. For this reason, in the literature have been proposed a great number of procedures and techniques in order to evaluate edge detectors. However, this continues being  an open problem, due to complexity of images.

The performance of our proposed algorithm is compared to the classical algorithms such as: Canny, Sobel, Prewitt and Roberts. A limitation of the classical methods for edge detection is that they operate, only, over the range domain, i.e., in the gray levels. Today, most of edge detectors process the image in the range-spatial domain, and the obtained results are better.

Our strategy works in the range-spatial domain and this guarantees the continuity of the edges. This does not happen with the classical methods. 

In the next section, we will carry out an experimental comparison between the detected edges with the classical algorithms and the detected edges  with our approach.

\section{Results and Validation}
In this section, we will show the obtained results of applying our edge detection algorithm. Our algorithm was applied to segmented images (6303, 41006 and 175083) of the Berkeley's database. The performance of the proposed method was tested and this was compared with other different approaches (classical edge detections).

Firstly, the obtained edges with the new proposed algorithm were compared with the accepted manual ground-truth segmentation of the Berkeley's database. One can note in Figures \ref{fig:experimentI}, \ref{fig:experimentII} and \ref{fig:experimentIII}, that there is not too much difference among the obtained edge images with the EDD algorithm and the true edge segmentation of the Berkeley's database.

A quantitative validation was carried out using several metrics: the probabilistic rank index (PRI), normalized probabilistic rank index (NPRI) and natural entropy distance (NED). The exposed results in Table \ref{tab:experiment_results} show that edge detection carried out with our EDD algorithm is highly competitive compared with other previously proposed algorithms \cite{Canny1986,Lakshmi2010,Papari2011,Michailovich2007}. In Table \ref{tab:experiment_comparation}, we can observe how the values of PRI, NPRI and NED of the resulting edge image of our algorithm are similar and sometimes greater, than the values of the resulting edge image of the classical algorithms.

In Figure \ref{fig:Comparation}, one can see the principal problem of the classical methods: the  edge discontinuities. However, our proposed strategy guarantees the edge continuity when operating in the spatial-range domain.

An example of an applications of our proposed strategy to a medical image is shown in Figure \ref{fig:Results}. In this case, we used different values of $h_s$ and $h_r$ in the $MSHi$. This is a preliminary result. A deeper paper about these results will be published.

\section{Conclusion}
In this work, we proposed a new edge detector algorithm based on the kernel density estimation. EDD introduces a new strategy for automatically find the threshold value based on the minimum of the density function. The proposed algorithm was applied to ground-truth of the Berkely's database and medical images.

Our method has been compared with the classical edge detection algorithms. The quality and quantity results are appropriate according to the criteria of specialists. Our method have a computational complexity similar to the classical edge detector with the advantage of preserving the edge continuity. 

The extensive experimental evaluation showed that our edge detection method is significantly a competitive algorithm. In future works, the experimental results related to a real problem in the field medical image will be expanded.


\ifCLASSOPTIONcaptionsoff
  \newpage
\fi
\bibliographystyle{IEEEtran}

\hfill

%
%
%



\begin{IEEEbiography}
[{\includegraphics[width=1in,height=1.4in]{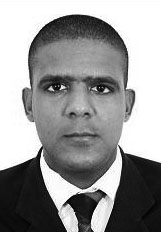}}]{Osvaldo Pereira.}
Received his bachelor degree in Computer Science Engineering in October 2008. He received his degree of Master of Science in mention of Applied Computer in October 2010. Now is part of the Digital Signal Processing Group of the Institute of Cybernetics, Mathematics and Physics (ICIMAF). His research interests include: processing and segmentation of digital images, reconstruction of three-dimensional models from medical imaging, visualization and virtual reality. He is developing his PhD in topics of edge detection in images.
\end{IEEEbiography}

\begin{IEEEbiography}[{\includegraphics[width=1in,height=1.3in]{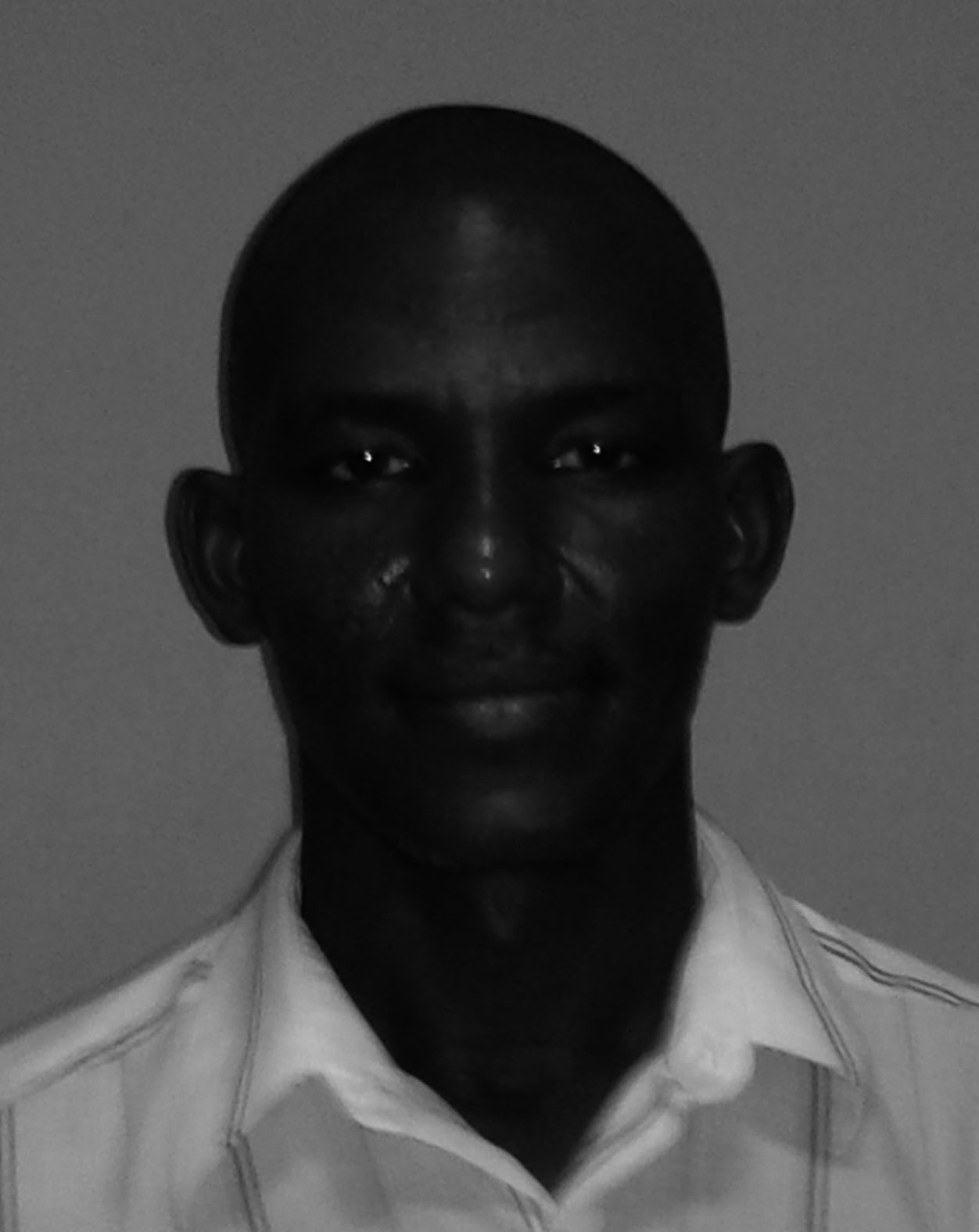}}]{Esley Torres.}
Received his bachelor degree in Mathematics from the Havana University in 2009. Since 2011, he is part of the Digital Signal Processing Group of the Institute of Cybernetics, Mathematics and Physics (ICIMAF). His research interests include segmentation, restoration, visual pattern recognition, and analysis of images. Since 2009, he teaches mathematics at Polytechnic University José Antonio Echavarría (ISPJAE). He has published more than 12  articles in international journals and has participated in many  international conferences. He has received one national prize.
\end{IEEEbiography}

\begin{IEEEbiography}[{\includegraphics[width=1in,height=1.3in]{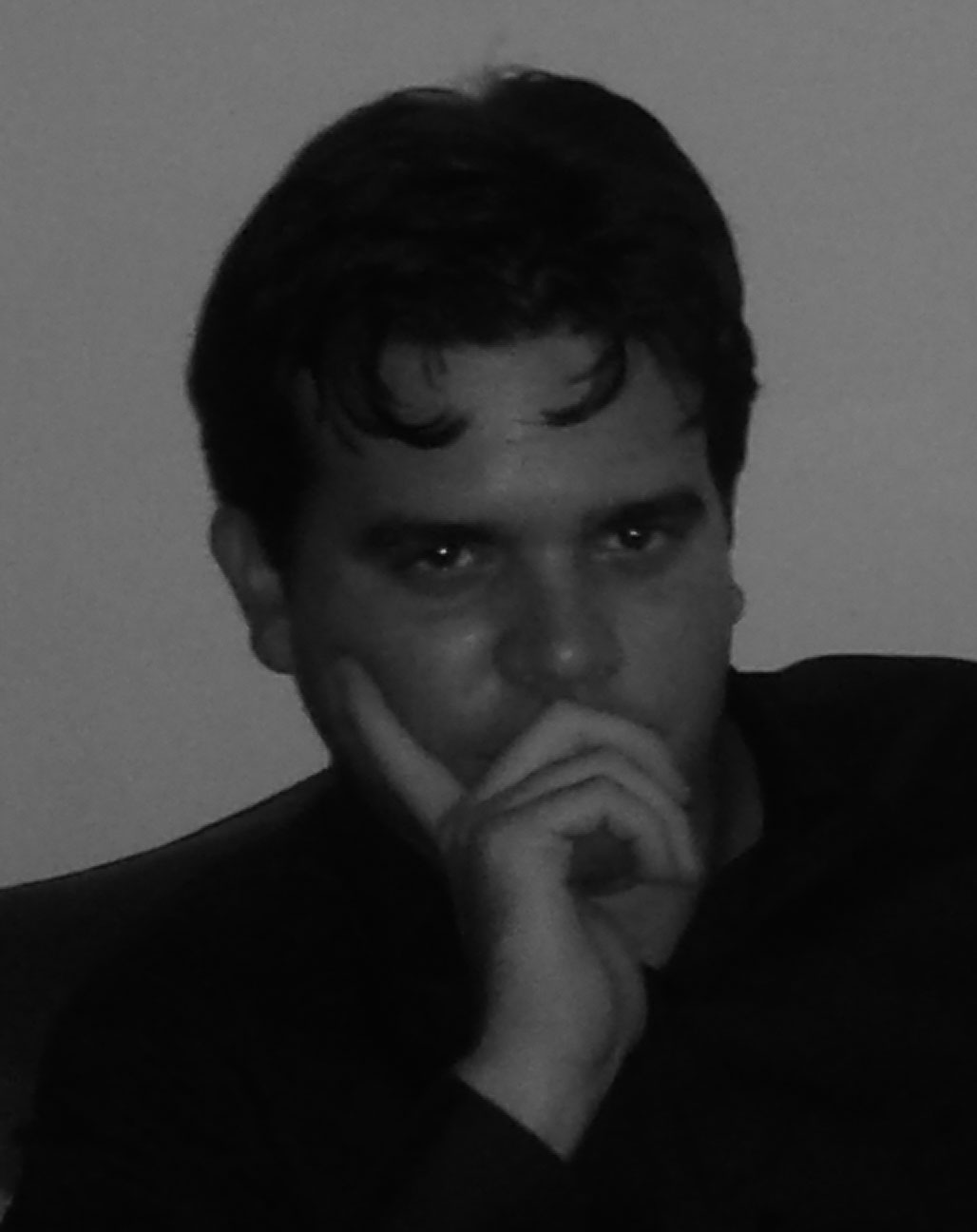}}]{Yaser Garc\'es.}
Received his bachelor degree in Mathematics from the Havana University in 2011 and the Master degree in 2014. Since 2011, he is part of the Digital Signal Processing Group of the Institute of Cybernetics, Mathematics and Physics (ICIMAF). His research interests include segmentation, restoration, visual pattern recognition, and analysis of images. He has published more than 10 articles in international journals and has participated in many international conferences. He has received more of five national prizes.
\end{IEEEbiography}

\begin{IEEEbiography}
[{\includegraphics[width=1in,height=1.3in]{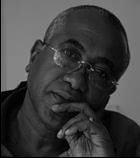}}]{Roberto Rodr\'iguez.}
Received his diploma in Physic from the Physics Faculty, Havana University in 1978 and the PhD degree from Technical University of Havana, in 1995. Since 1998, he is the head of the Digital Signal Processing Group of the Institute of Cybernetics, Mathematics and Physics (ICIMAF). His research interests include Segmentation, Restoration, Mathematical Morphology, Visual pattern recognition, Analysis and Interpretation of images, Theoretical studies of Gaussian Scale-Space and Mean shift. He has published more than 100 articles in international journals and in many international conferences. He has two published books and three written chapters in other related books with the speciality. He has received more of ten prizes national e international.
\end{IEEEbiography}

\clearpage
\onecolumn
\vspace{-.2in}
\section*{Appendix}
\vspace{-.1in}
\begin{figure}[ht]
\centering
\begin{tabular}{ccc}
Image 6303 of Berkeley's Database & Image 41006 of Berkeley's Database & Image 175083 of Berkeley's Database\\ 
\includegraphics[width=0.2\paperwidth]		{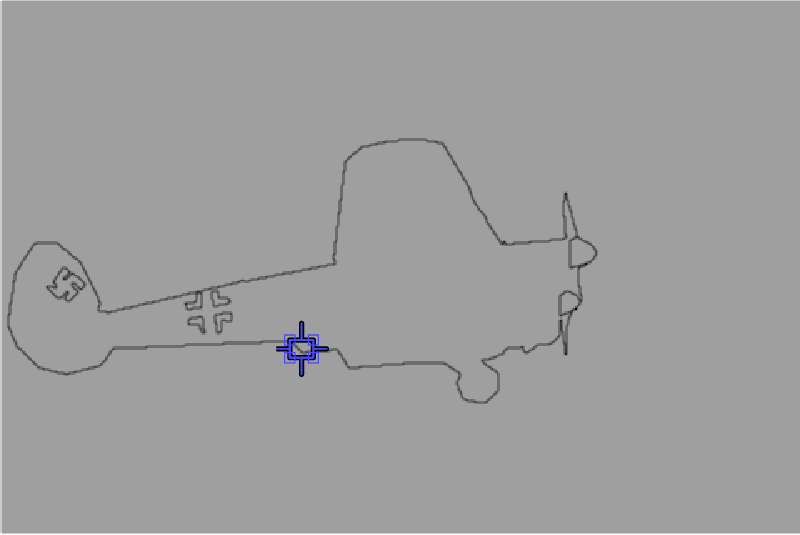}\label{fig:3063_u} &
\includegraphics[width=0.2\paperwidth]		{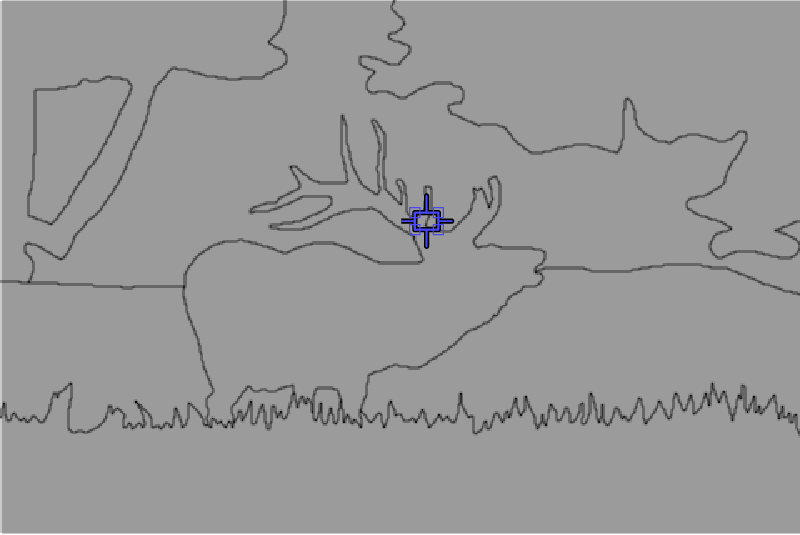}\label{fig:41006_u} &
\includegraphics[width=0.2\paperwidth]		{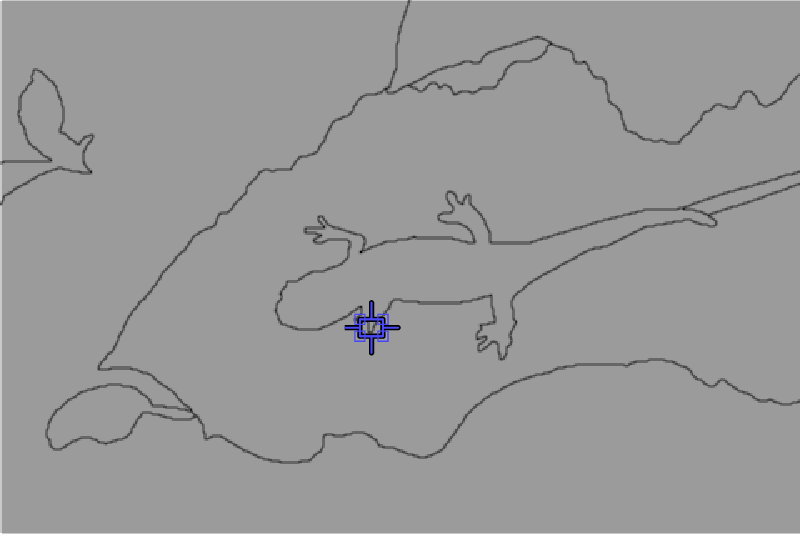}\label{fig:175083_u} \\ 

\includegraphics[width=0.2\paperwidth,  height=0.145\paperwidth]{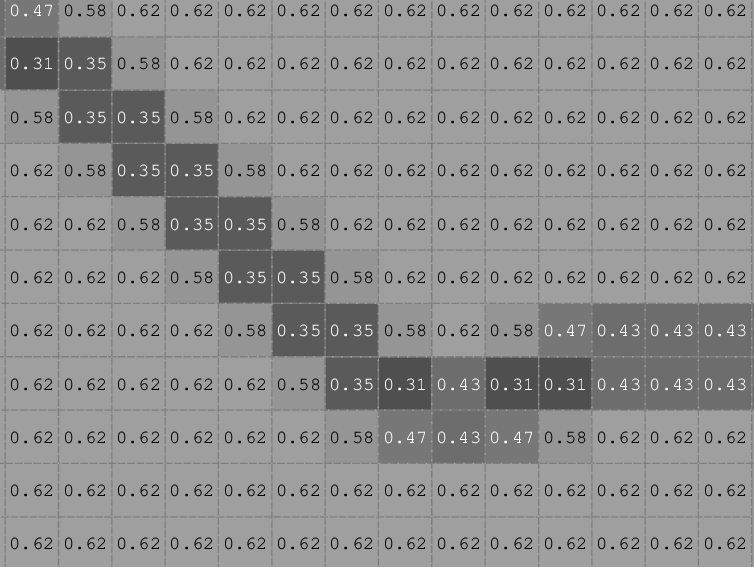}\label{fig:3063_part} &  \includegraphics[width=0.2\paperwidth]{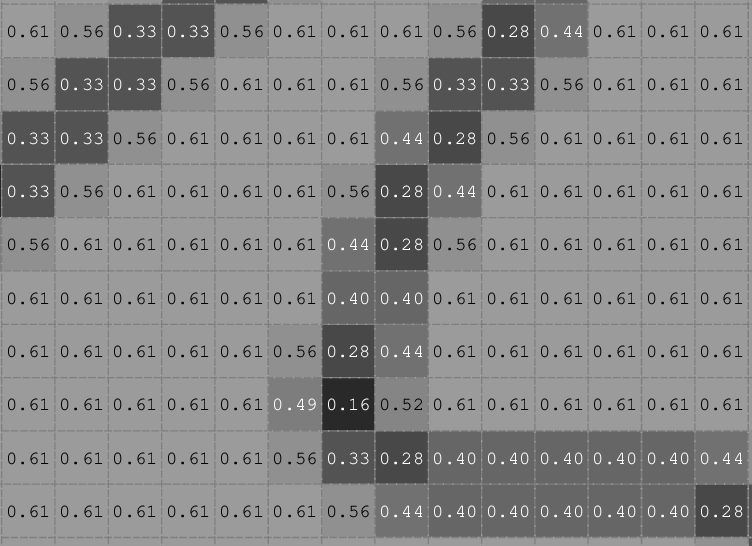}\label{fig:41006_part} & 
 \includegraphics[width=0.2\paperwidth]{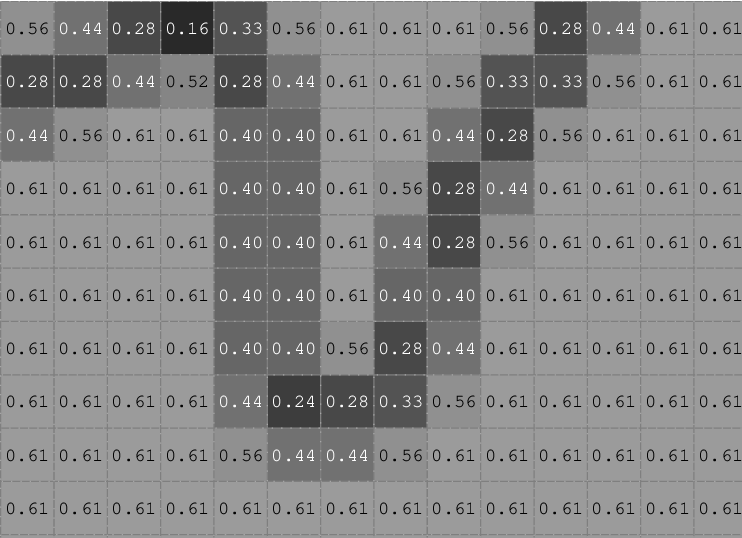}\label{fig:175083_part} \\

\includegraphics[width=0.23\paperwidth]{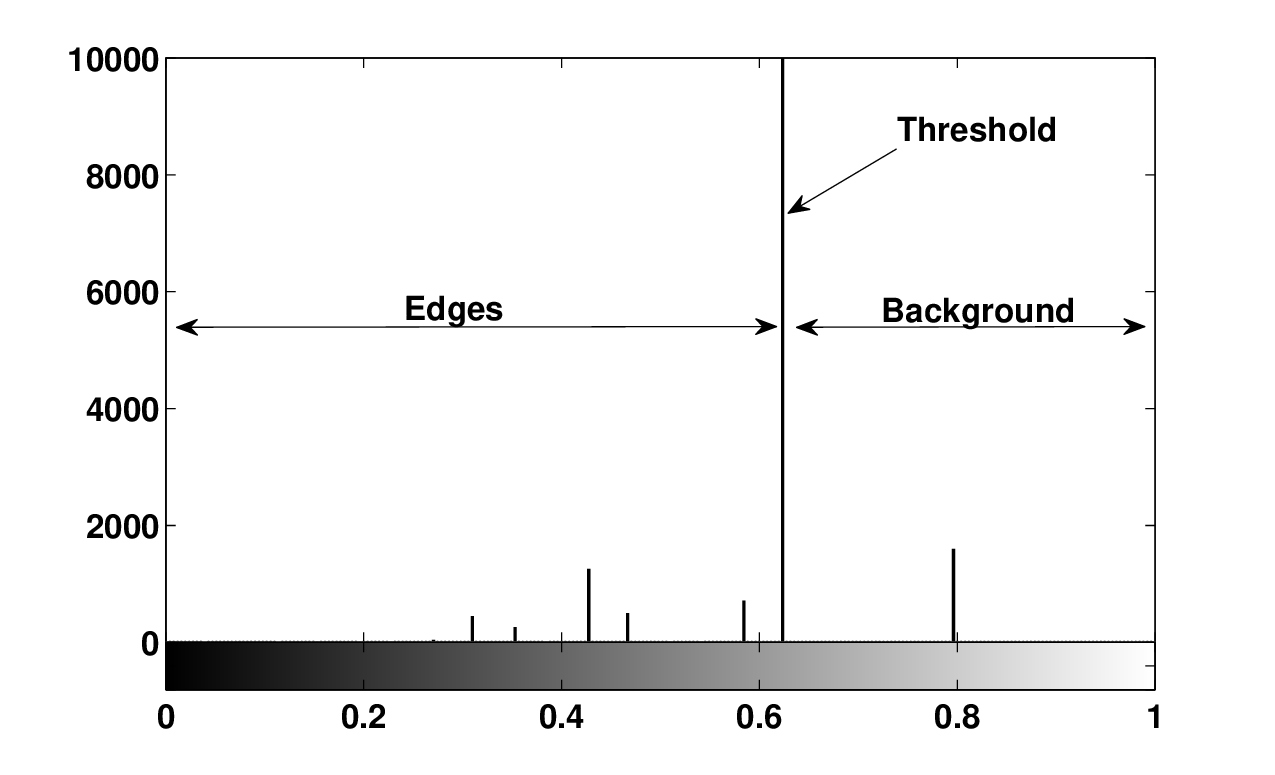}\label{fig:3063_hist} &
\includegraphics[width=0.23\paperwidth]{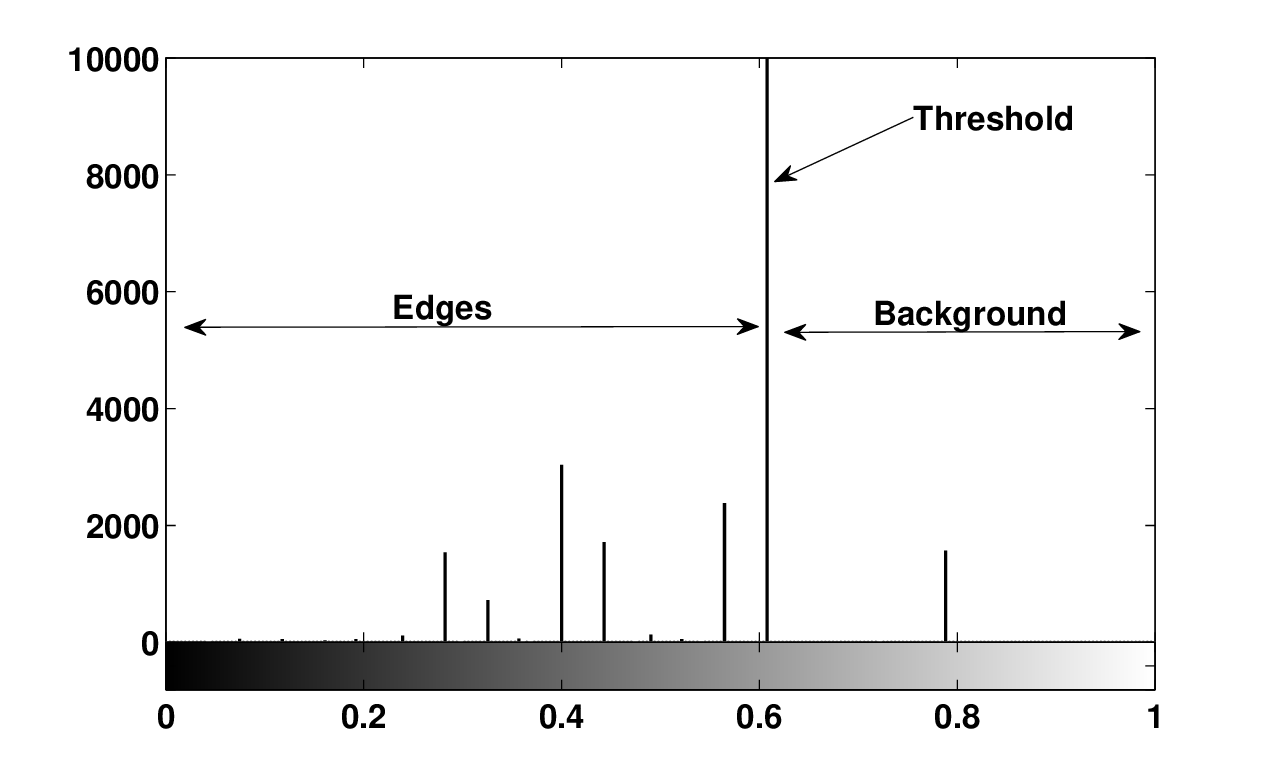}\label{fig:41006_hist} &
 \includegraphics[width=0.23\paperwidth]{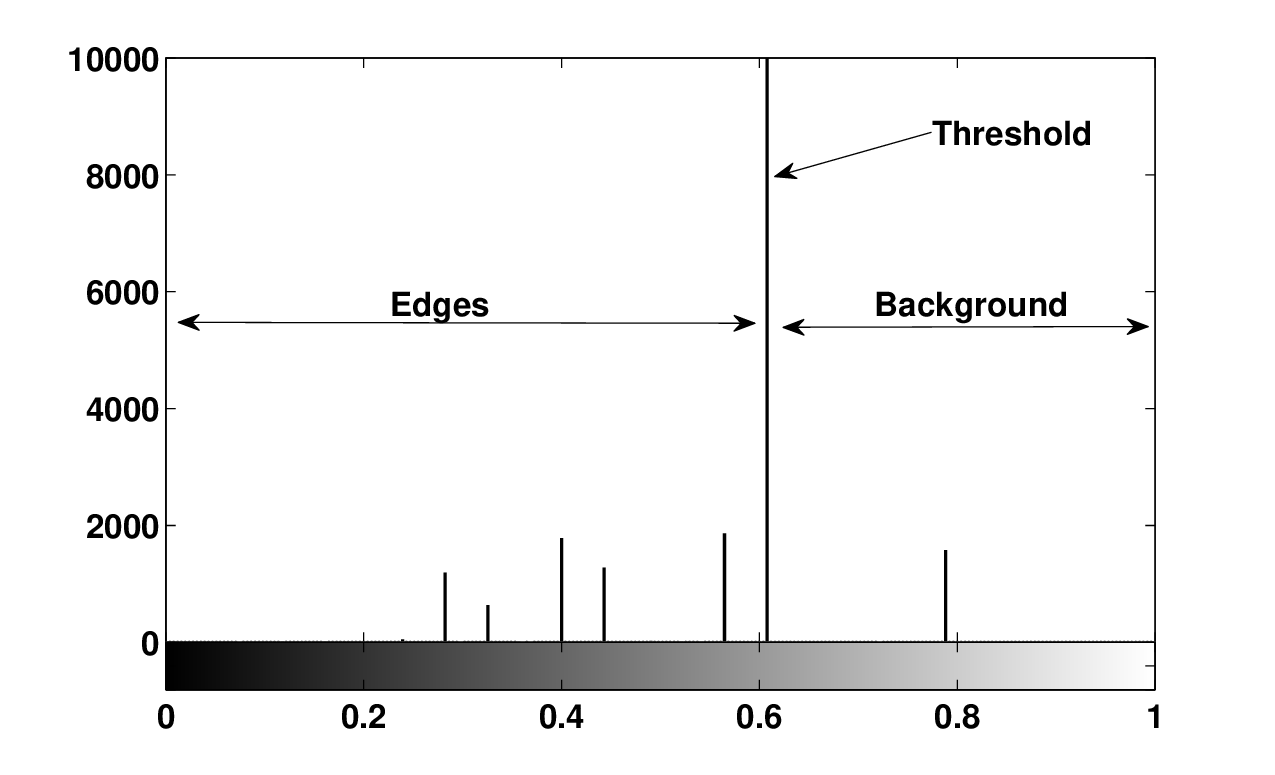}\label{fig:175083_hist} \\
\end{tabular}
\caption{Automatic calculation of the threshold value. 
$1^{st}$ Row: Density images obtained with EDD algorithm. $2^{nd}$ Row: Zoom of the region in the density images. $3^{th}$ Row: Histograms of frecuency of the density images.}
\label{fig:experiment_threshold}
\end{figure}

\begin{figure}[ht]
\centering
\subfigure[]{\includegraphics[scale=0.398]{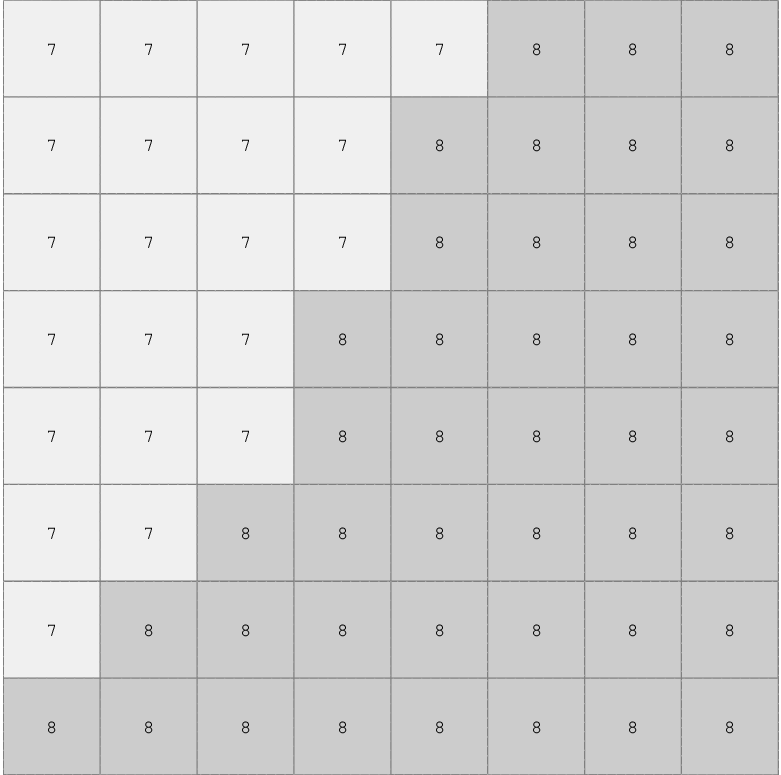}}
\subfigure[]{\includegraphics[scale=0.6]{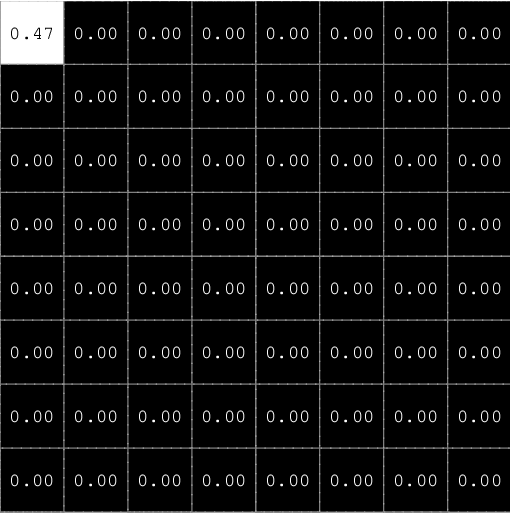}} 
\subfigure[]{\includegraphics[scale=0.6]{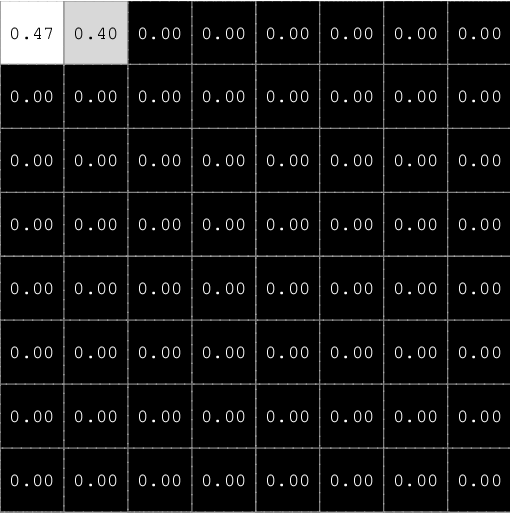}} 
 \\\vspace{-.05in}
\subfigure[]{\includegraphics[scale=0.6]{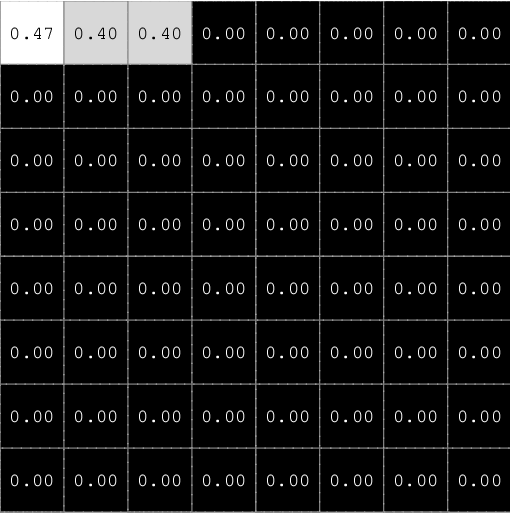}}
\subfigure[]{\includegraphics[scale=0.6]{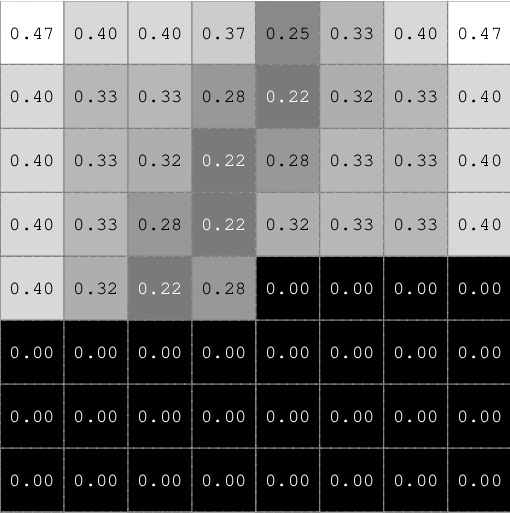}} 
\subfigure[]{\includegraphics[scale=0.6]{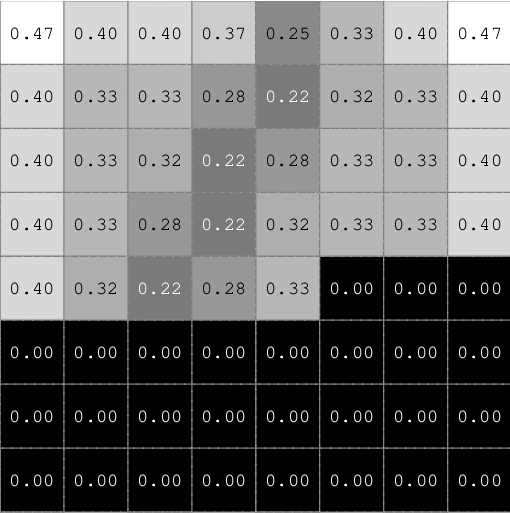}} 
\\\vspace{-.05in}
\subfigure[]{\includegraphics[scale=0.6]{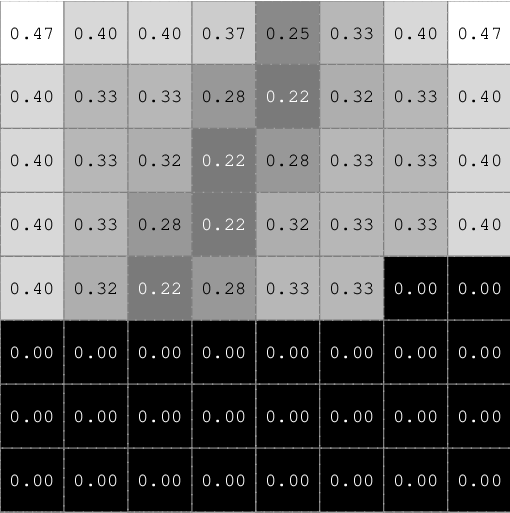}} 
\subfigure[]{\includegraphics[scale=0.6]{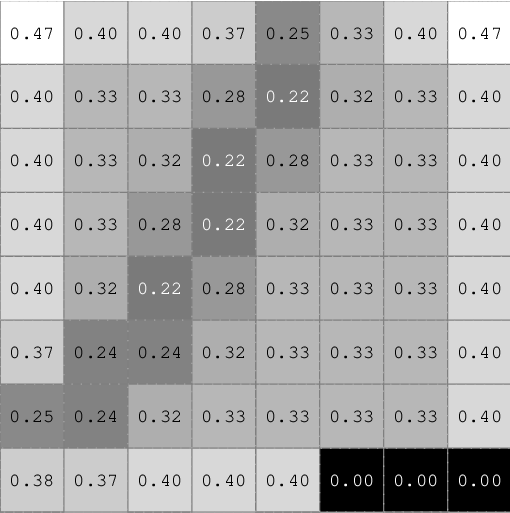}} \subfigure[]{\includegraphics[scale=0.6]{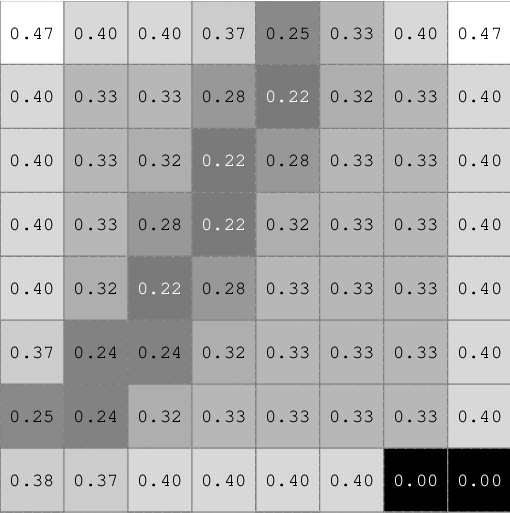}} 
\\\vspace{-.05in}
\subfigure[]{\includegraphics[scale=0.6]{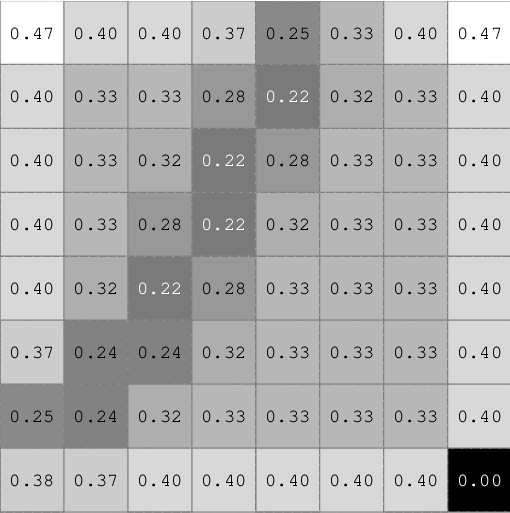}} 
\subfigure[]{\includegraphics[scale=0.6]{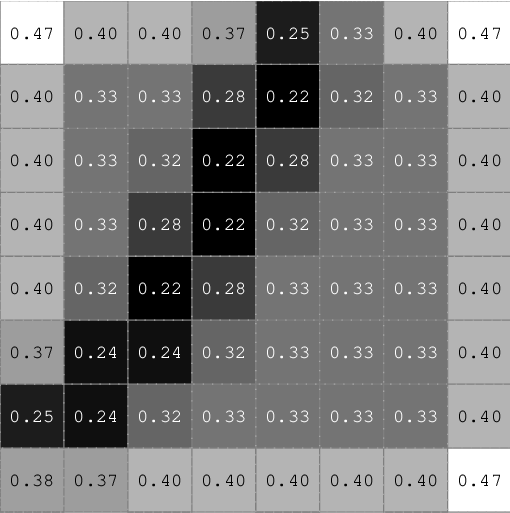}} 
\subfigure[]{\includegraphics[scale=0.600]{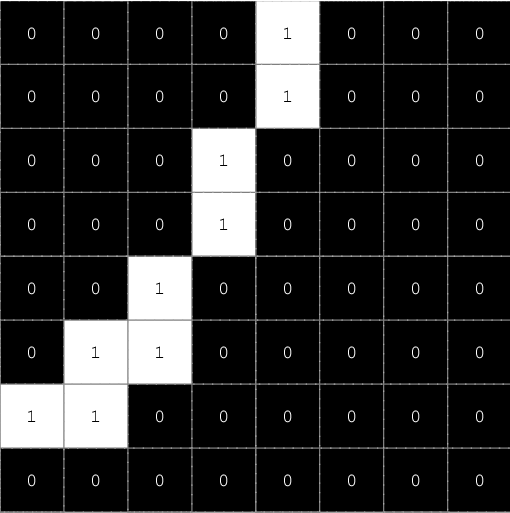}} \vspace{-.1in}
\caption{Algorithm Iterations. (a) Original image   (b)-(k) Algorithm iterations (l) Edges detected.}
\label{fig:AlgIterations}
\vspace{-.1in}
\end{figure}

\begin{figure}[ht]
\centering
\begin{tabular}{ccc}
\multicolumn{3}{c}{Experiment I: Image 6303 of Berkeley's Database.}\\
& Ground-Truth I & \\ 
\includegraphics[width=0.23\paperwidth]		{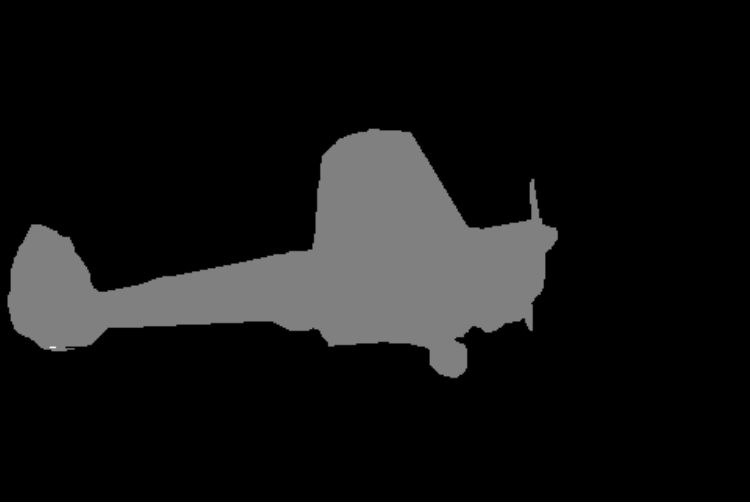}\label{fig:3063_1} & \includegraphics[width=0.23\paperwidth]{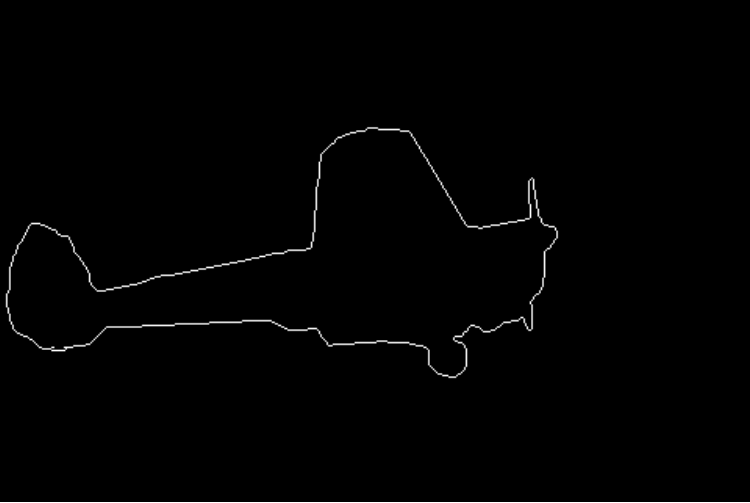}\label{fig:3063_b1} & \includegraphics[width=0.23\paperwidth]{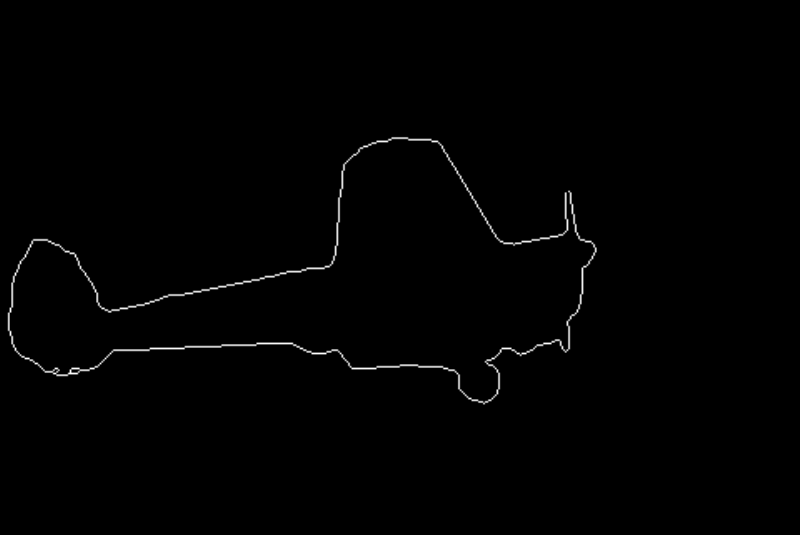}\label{fig:3063_a1} \\
& Ground-Truth II & \\ 
\includegraphics[width=0.23\paperwidth]{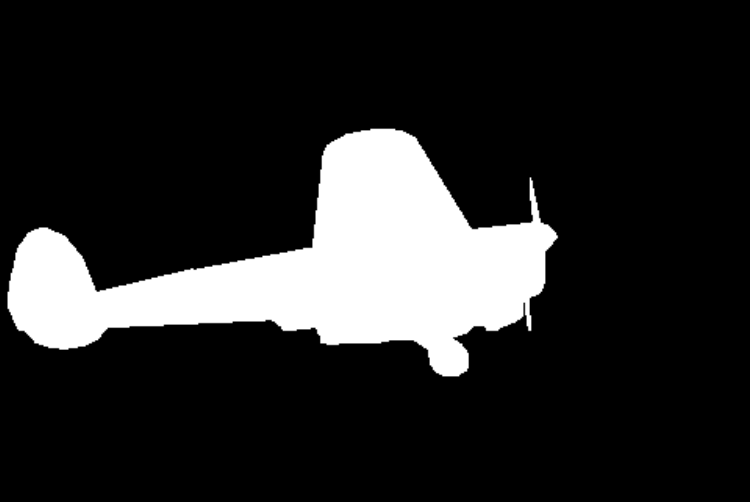}\label{fig:3063_2} & \includegraphics[width=0.23\paperwidth]{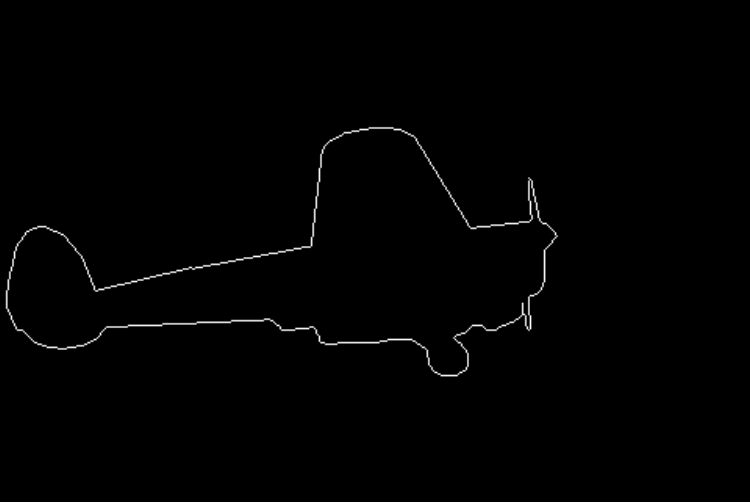}\label{fig:3063_b2} & \includegraphics[width=0.23\paperwidth]{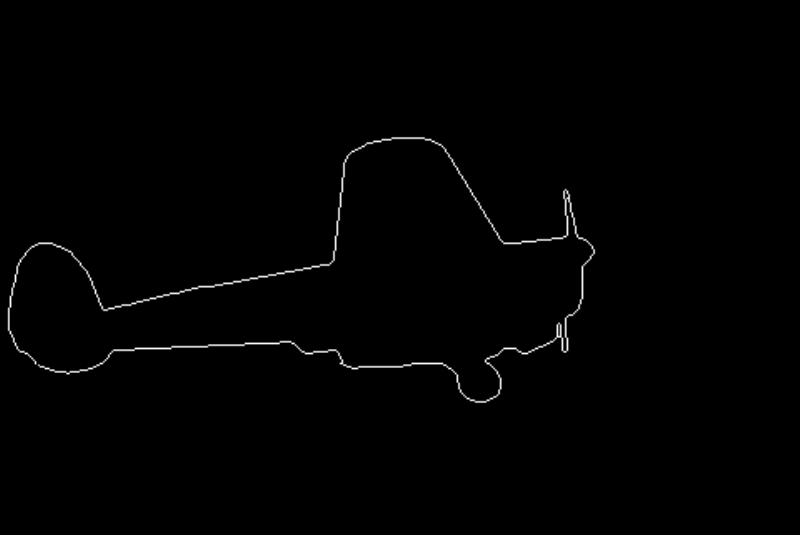}\label{fig:3063_a2} \\
& Ground-Truth III & \\ 
\includegraphics[width=0.23\paperwidth]{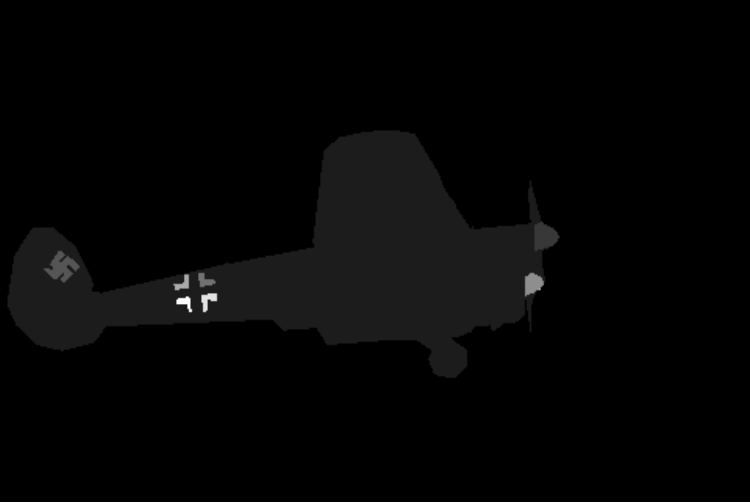}\label{fig:3063_4} & \includegraphics[width=0.23\paperwidth]{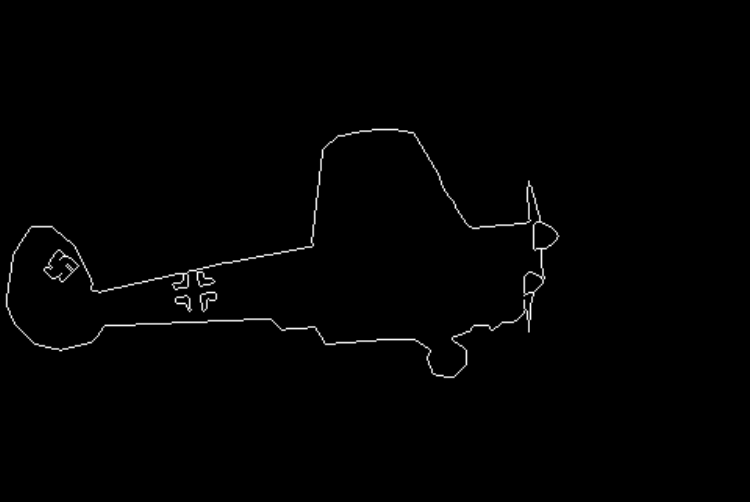}\label{fig:3063_b4} & \includegraphics[width=0.23\paperwidth]{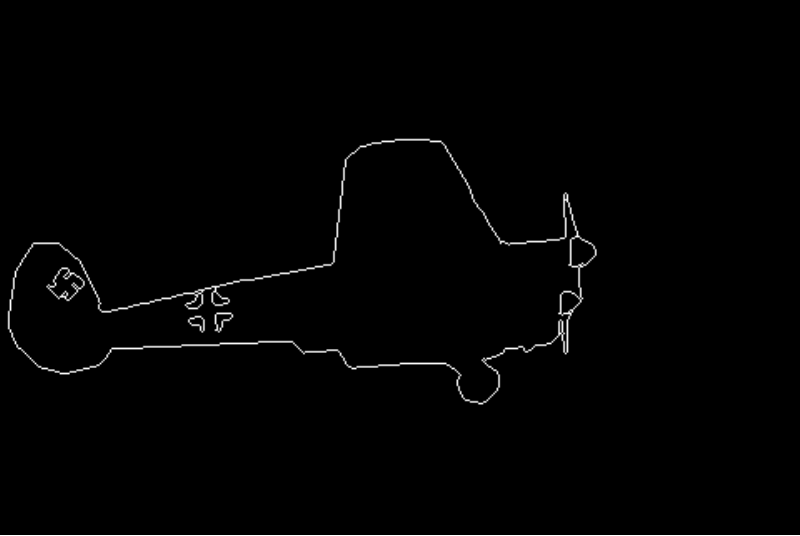}\label{fig:3063_a4} \\
& Ground-Truth IV & \\ 
\includegraphics[width=0.23\paperwidth]{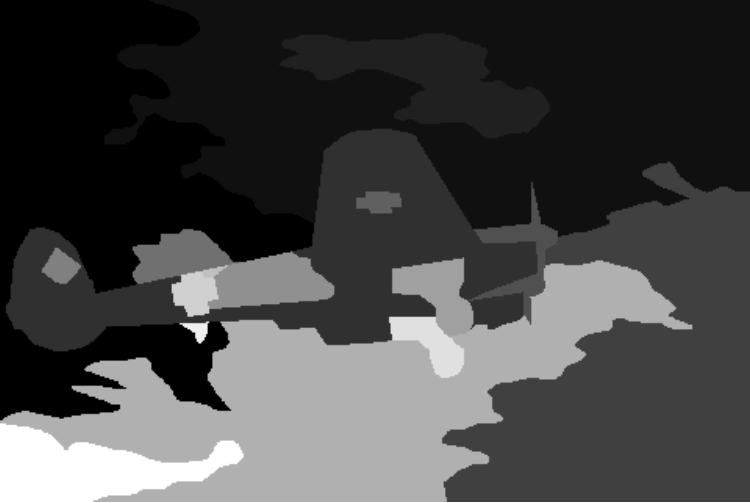}\label{fig:3063_5} & \includegraphics[width=0.23\paperwidth]{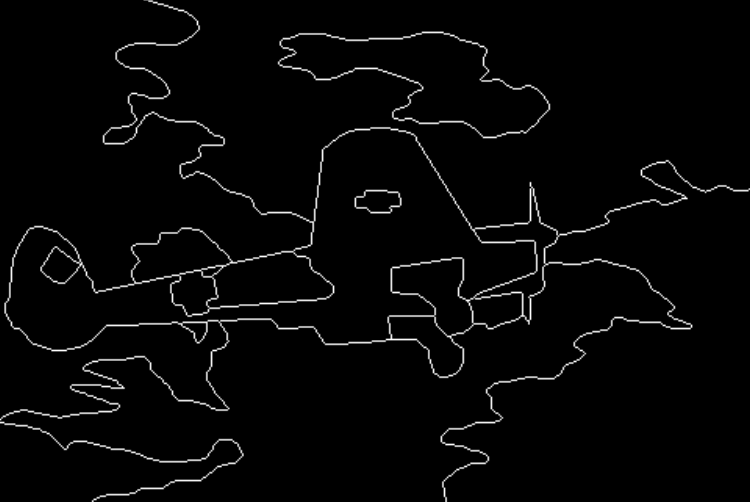}\label{fig:3063_b5} & \includegraphics[width=0.23\paperwidth]{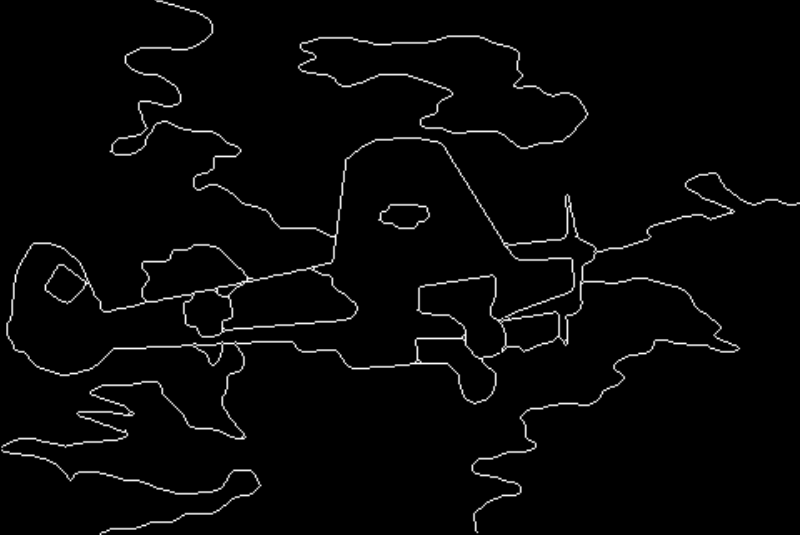}\label{fig:3063_a5} \\
& Ground-Truth V & \\
\includegraphics[width=0.23\paperwidth]{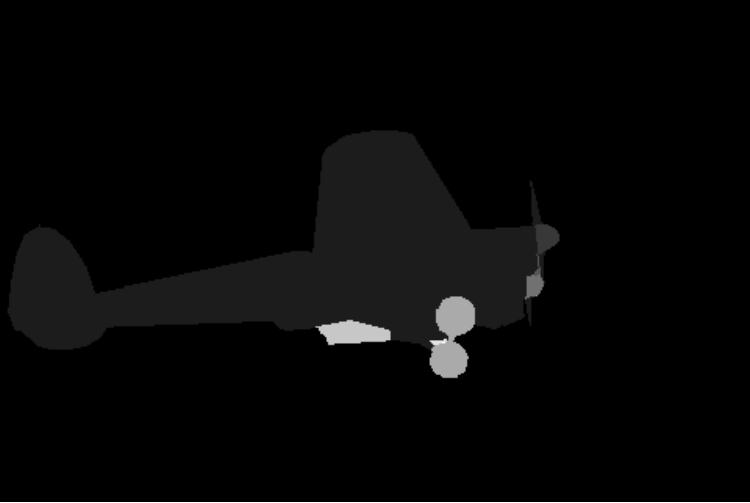}\label{fig:3063_6} & \includegraphics[width=0.23\paperwidth]{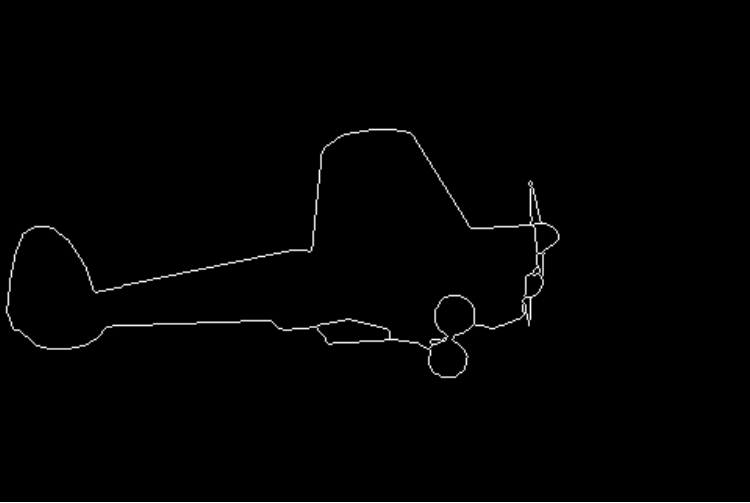}\label{fig:3063_b6} & \includegraphics[width=0.23\paperwidth]{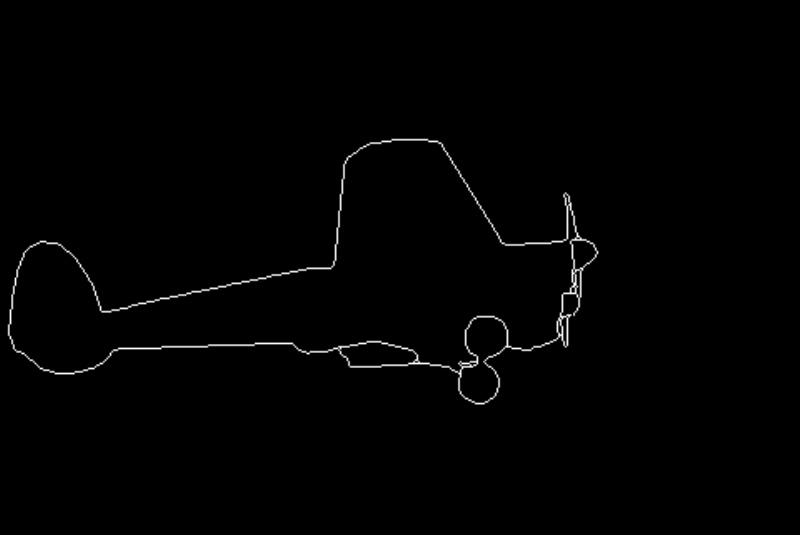}\label{fig:3063_a6} \\
\end{tabular}
\caption{Comparation of edge ground-truth segmentation with density algorithm segmentation. 
$1^{st}$ Column: Region ground-truth segmentation. $2^{nd}$ Column: Edge ground-truth segmentation. $3^{th}$ Column: Edge Detection by Density (EDD).}
\label{fig:experimentI}
\end{figure}

\begin{figure}[ht]
\centering
\begin{tabular}{ccc}
\multicolumn{3}{c}{Experiment II: Image 41006 of Berkeley's Database.}\\
& Ground-Truth I & \\
\includegraphics[width=0.23\paperwidth]		{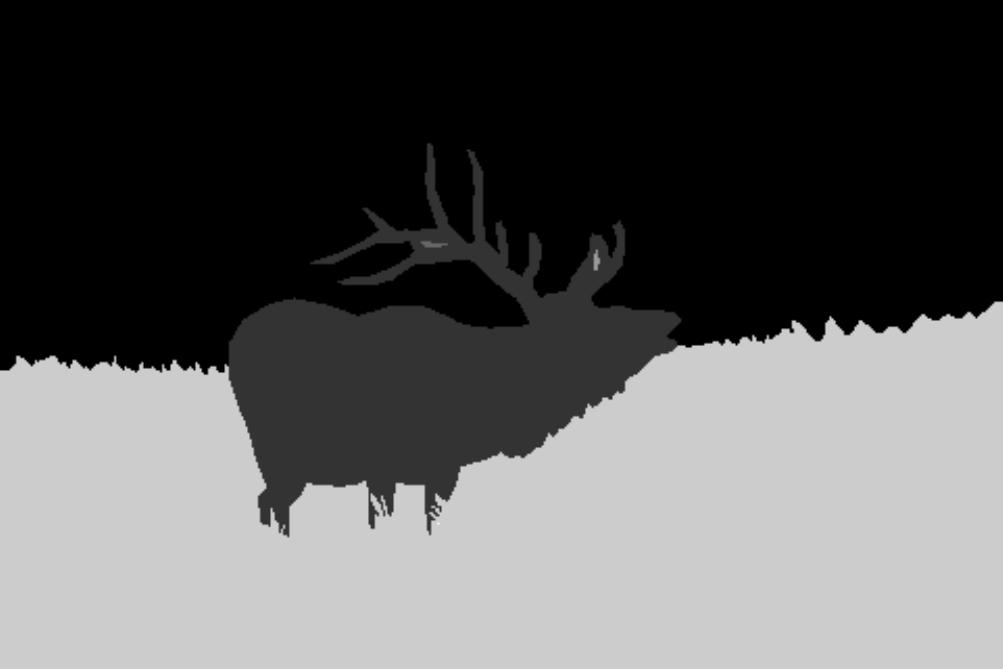} \label{fig:41006_1} & \includegraphics[width=0.23\paperwidth]	{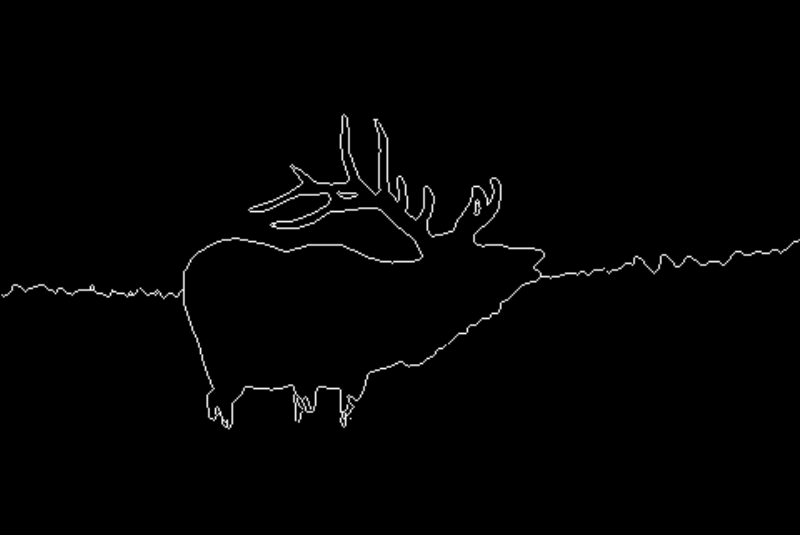} \label{fig:41006_a1} & \includegraphics[width=0.23\paperwidth]		{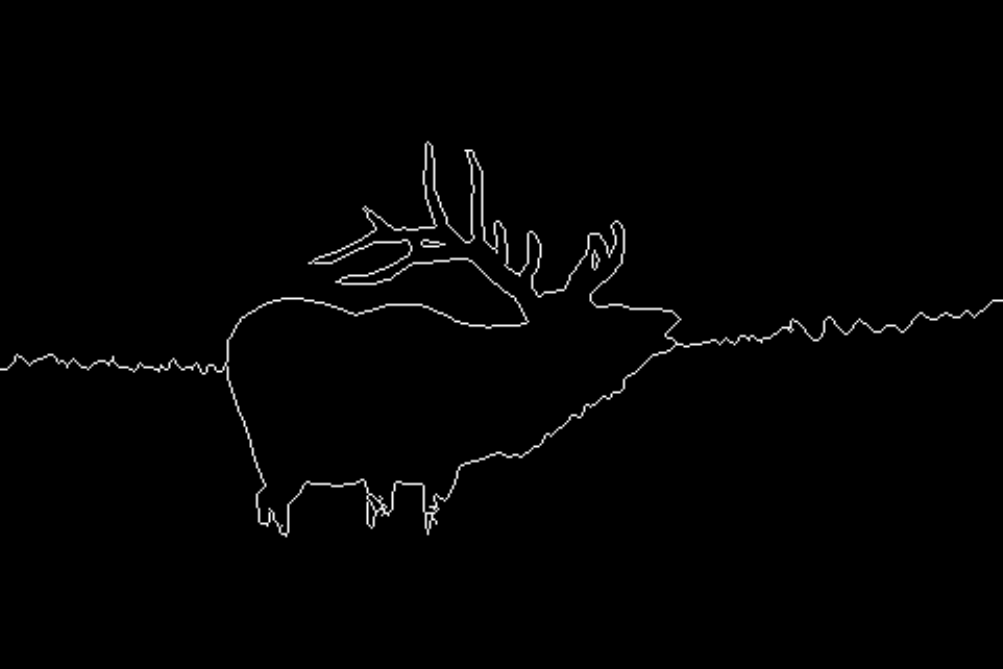} \label{fig:41006_b1} \\
& Ground-Truth II & \\
\includegraphics[width=0.23\paperwidth]		{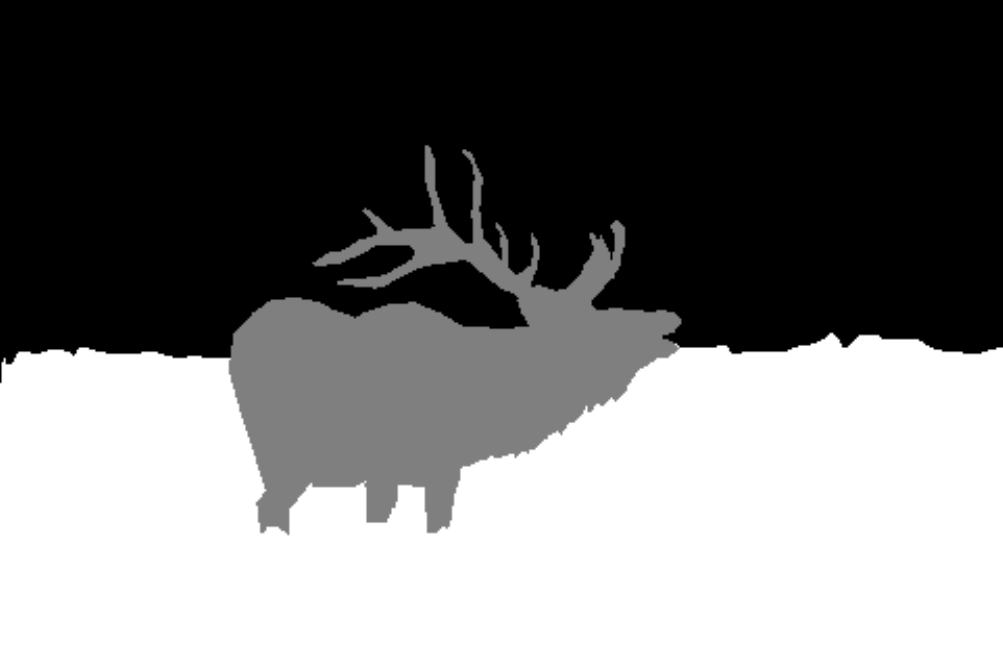} \label{fig:41006_2} & \includegraphics[width=0.23\paperwidth]		{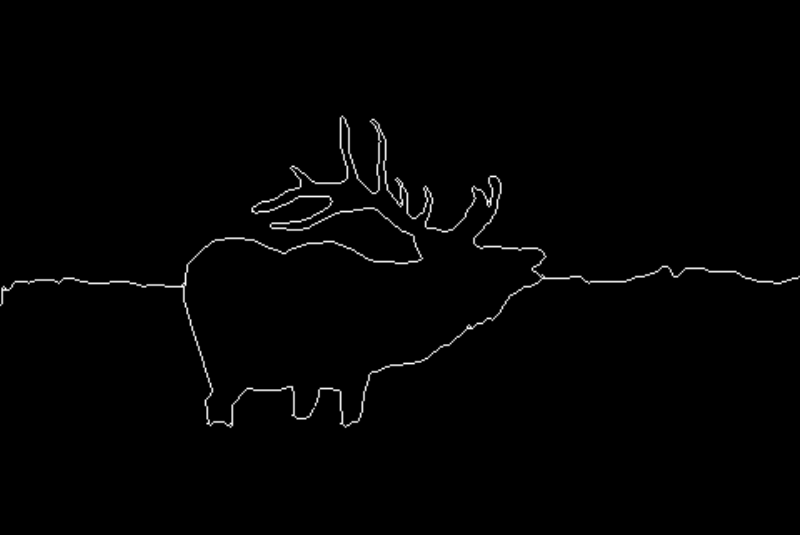} \label{fig:41006_a2} & \includegraphics[width=0.23\paperwidth]		{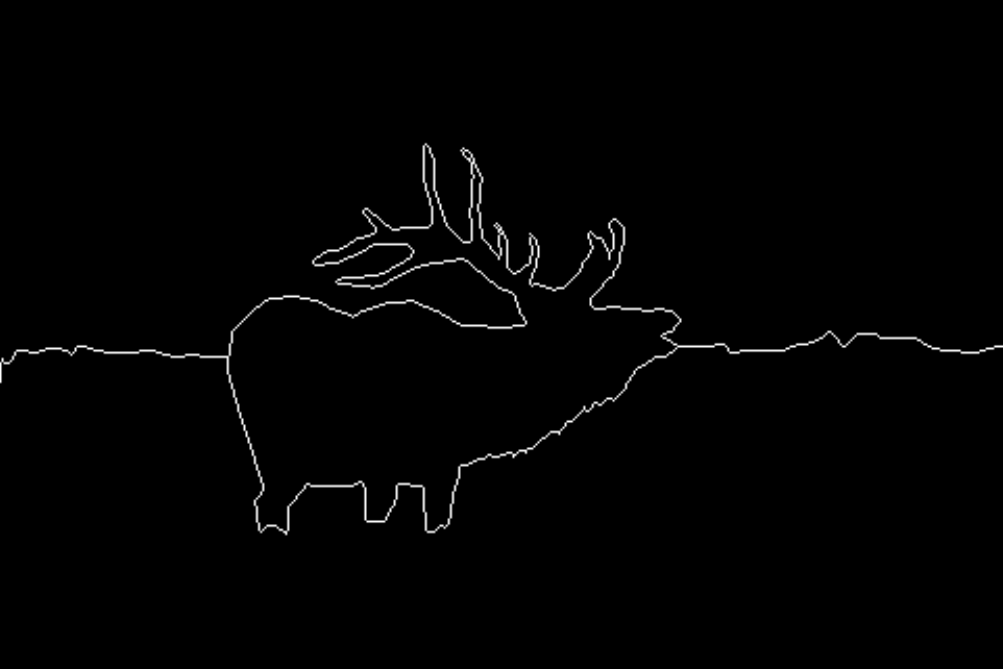} \label{fig:41006_b2}\\
& Ground-Truth III & \\
\includegraphics[width=0.23\paperwidth]		{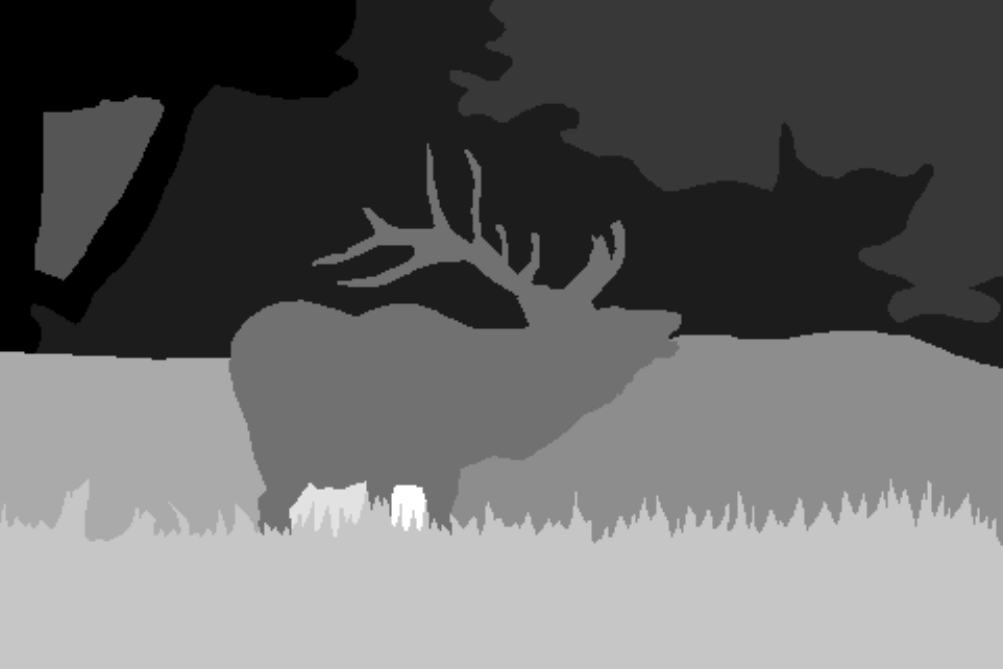} \label{fig:41006_3} & \includegraphics[width=0.23\paperwidth]		{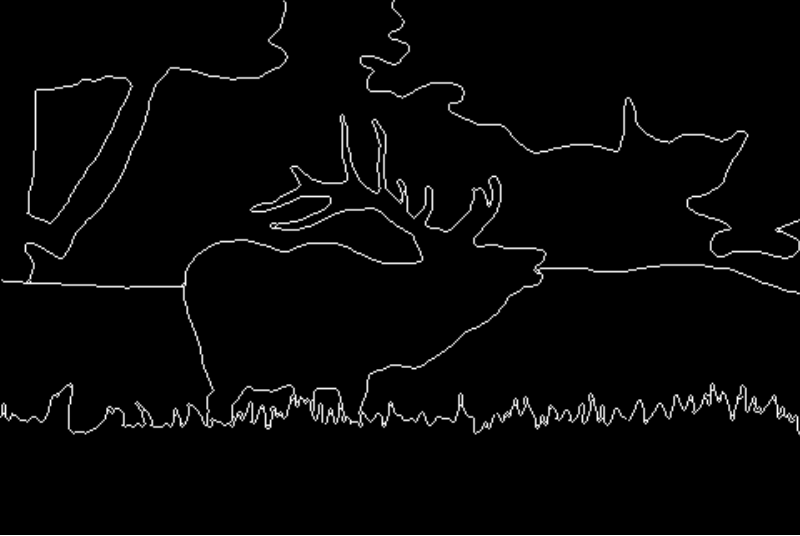} \label{fig:41006_a3} & \includegraphics[width=0.23\paperwidth]		{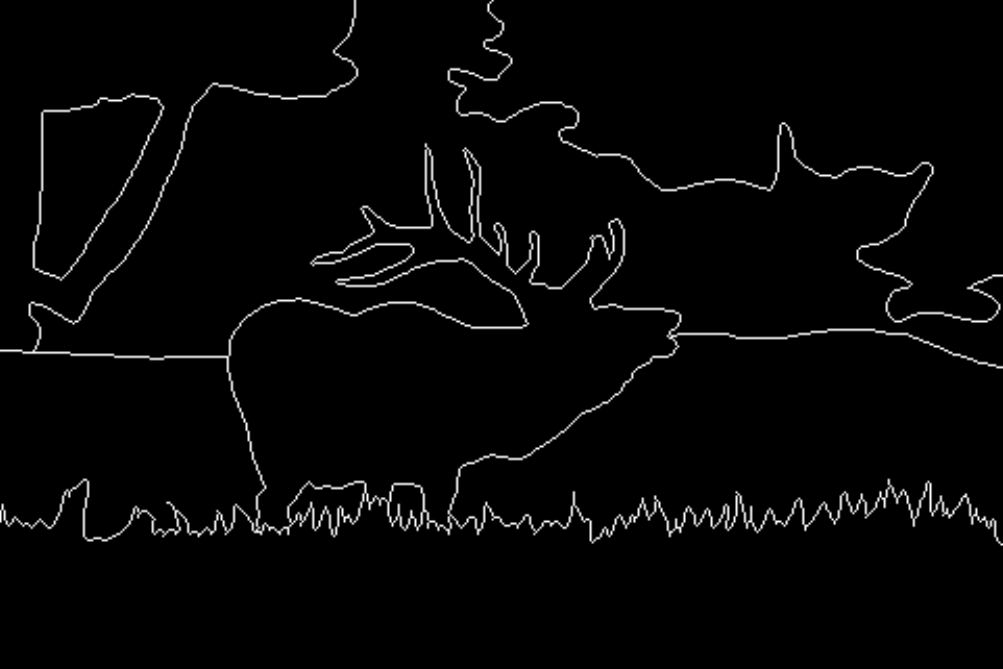} \label{fig:41006_b3} \\
& Ground-Truth IV & \\
\includegraphics[width=0.23\paperwidth]		{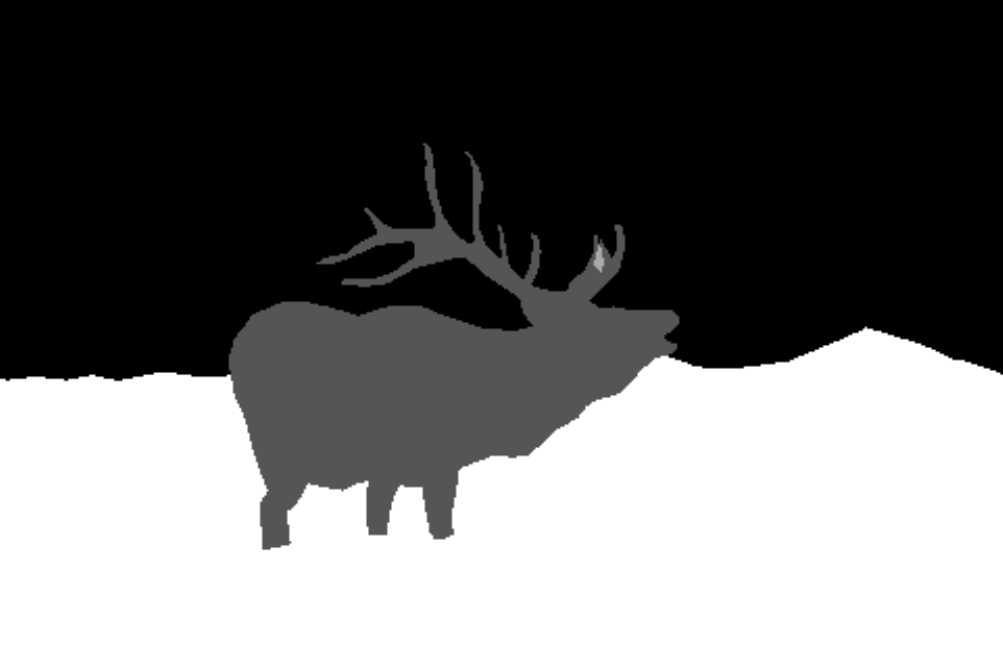}	\label{fig:41006_4} & \includegraphics[width=0.23\paperwidth]		{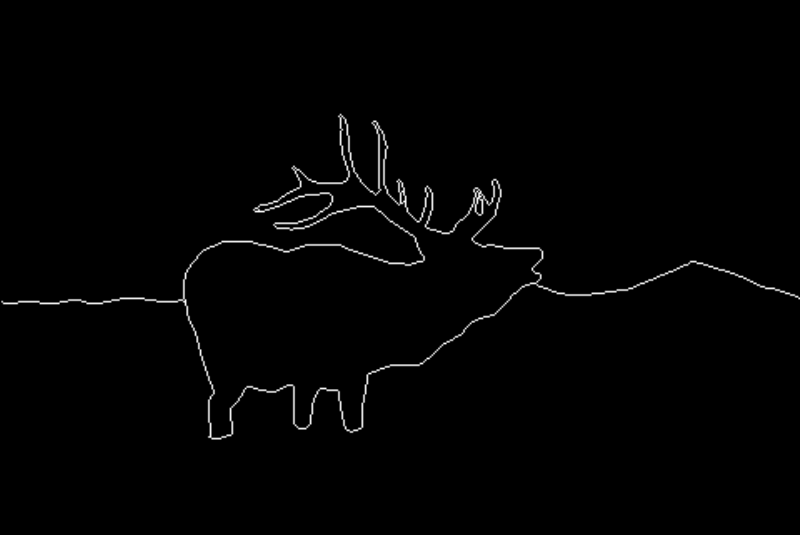} \label{fig:41006_a4} & \includegraphics[width=0.23\paperwidth]		{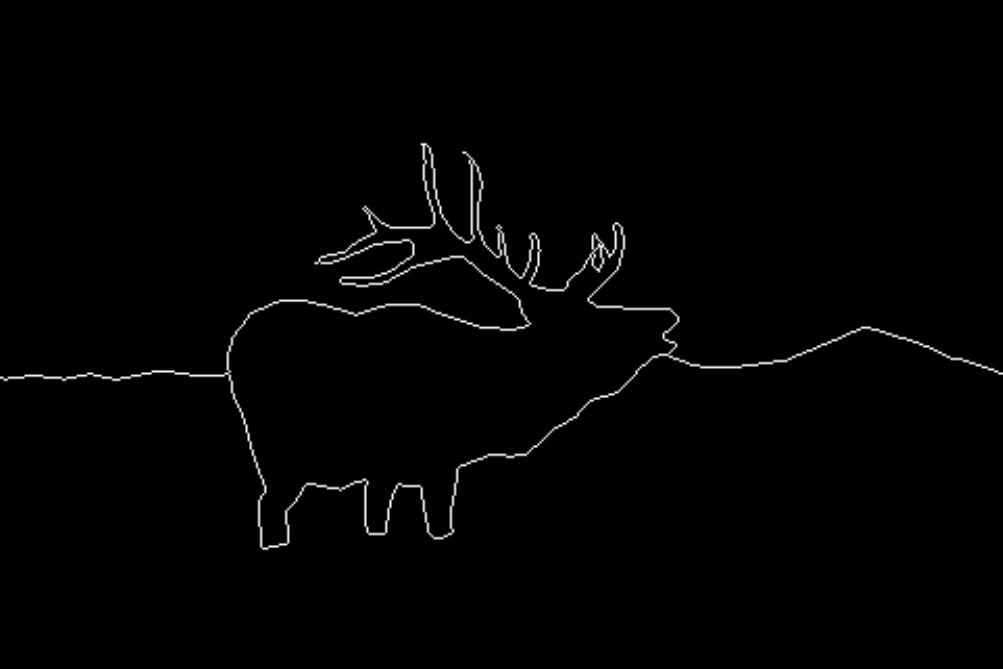} \label{fig:41006_b4} \\
& Ground-Truth V & \\
\includegraphics[width=0.23\paperwidth]		{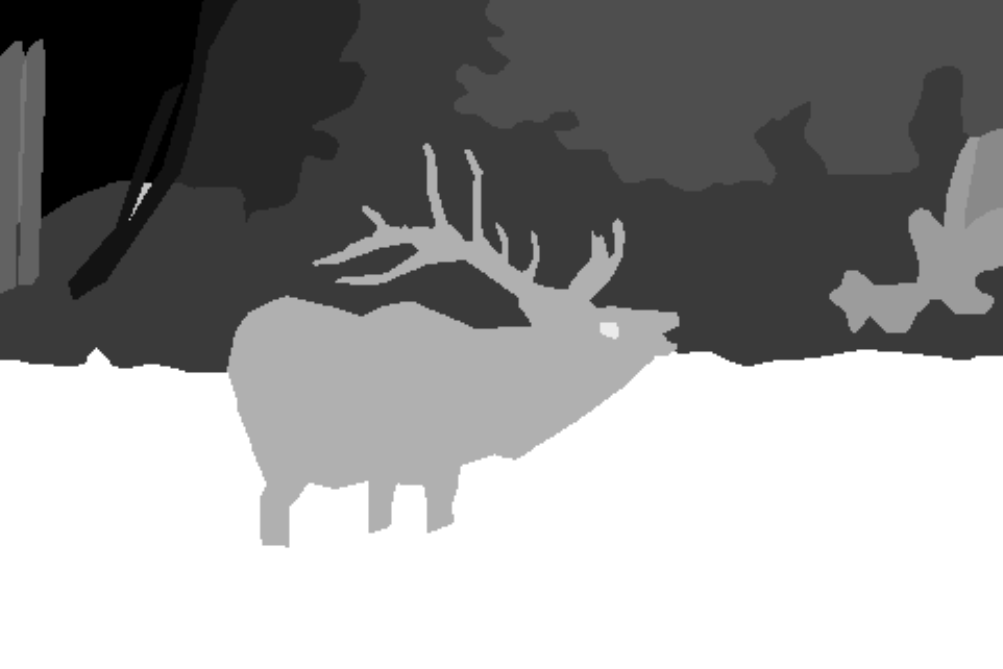}	\label{fig:41006_5} & \includegraphics[width=0.23\paperwidth]		{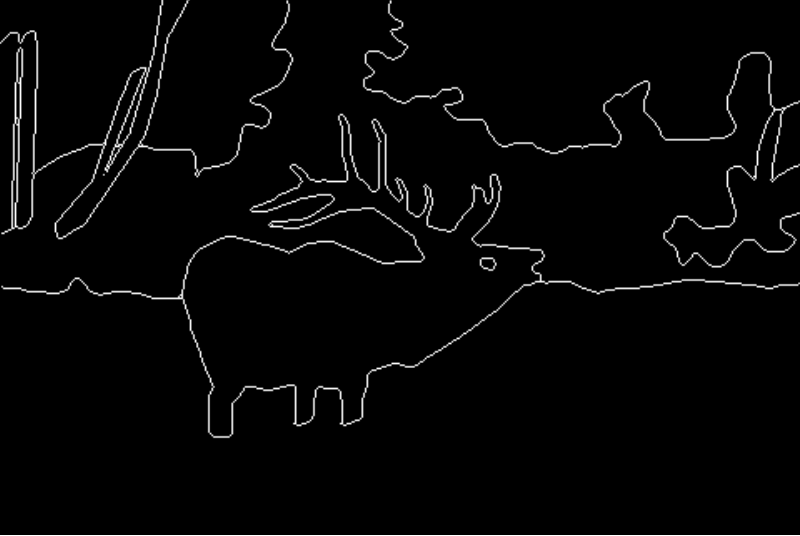} \label{fig:41006_a5} & \includegraphics[width=0.23\paperwidth]		{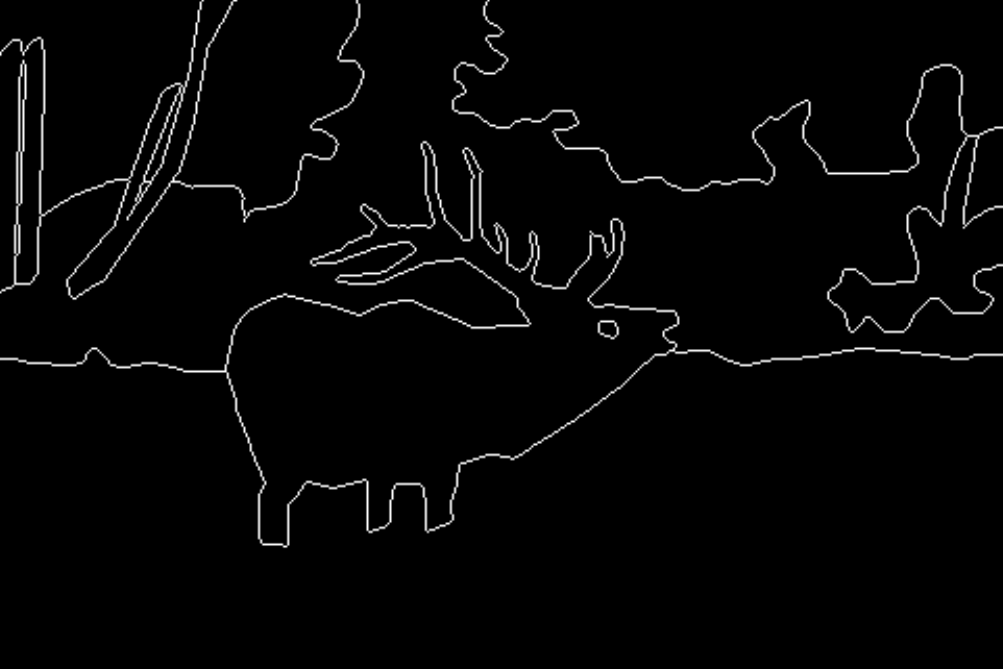} \label{fig:41006_b5} \\
\end{tabular}
\caption{Comparation of edge ground-truth segmentation with density algorithm segmentation. 
$1^{st}$ Column: Region ground-truth segmentation. $2^{nd}$ Column: Edge ground-truth segmentation. $3^{th}$ Column: Edge Detection by Density (EDD).}
\label{fig:experimentII}
\end{figure}

\begin{figure}[ht]
\centering
\begin{tabular}{ccc}
\multicolumn{3}{c}{Experiment III: Image 175083 of Berkeley's Database.}\\
& Ground-Truth I & \\
\includegraphics[width=0.23\paperwidth]		{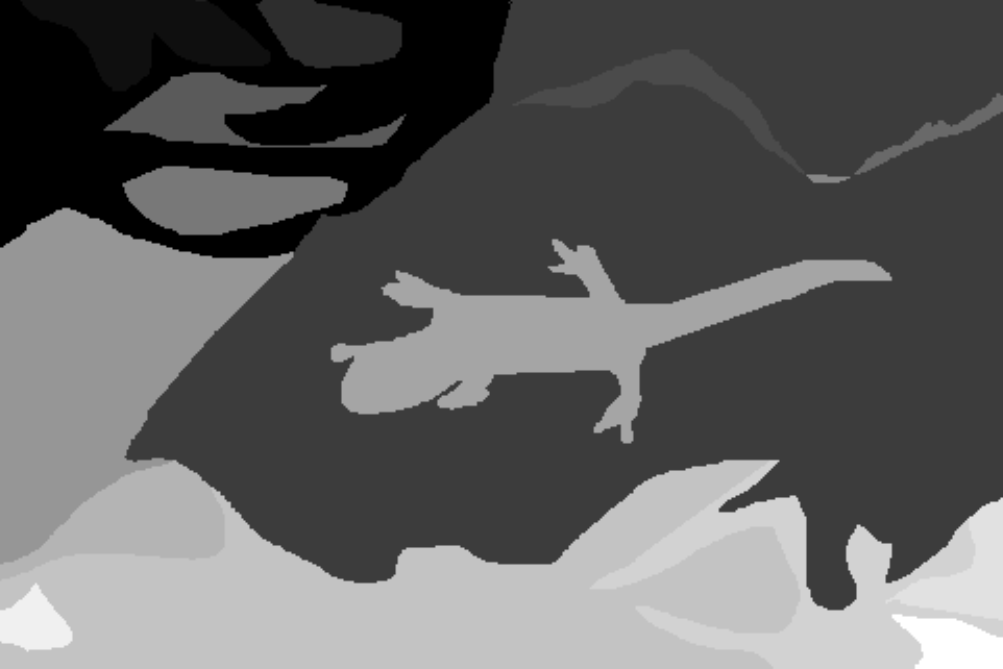} \label{fig:175083_1} &
\includegraphics[width=0.23\paperwidth]		{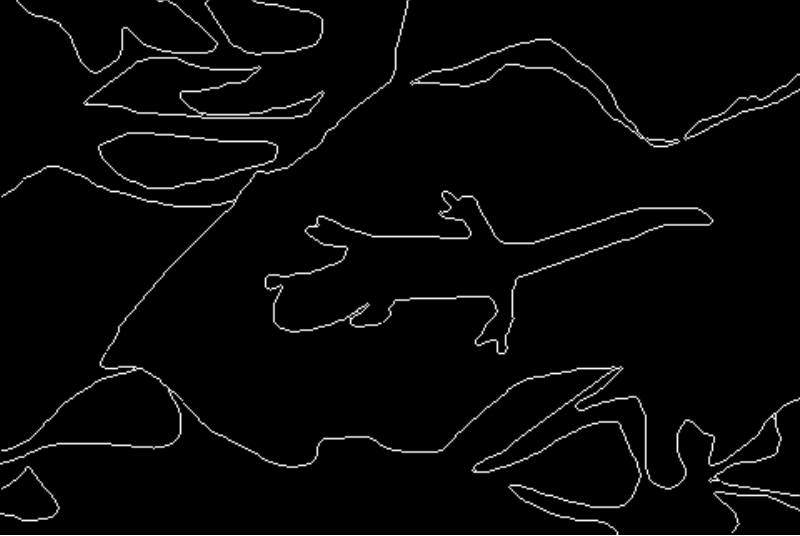} \label{fig:175083_a1} &
\includegraphics[width=0.23\paperwidth]		{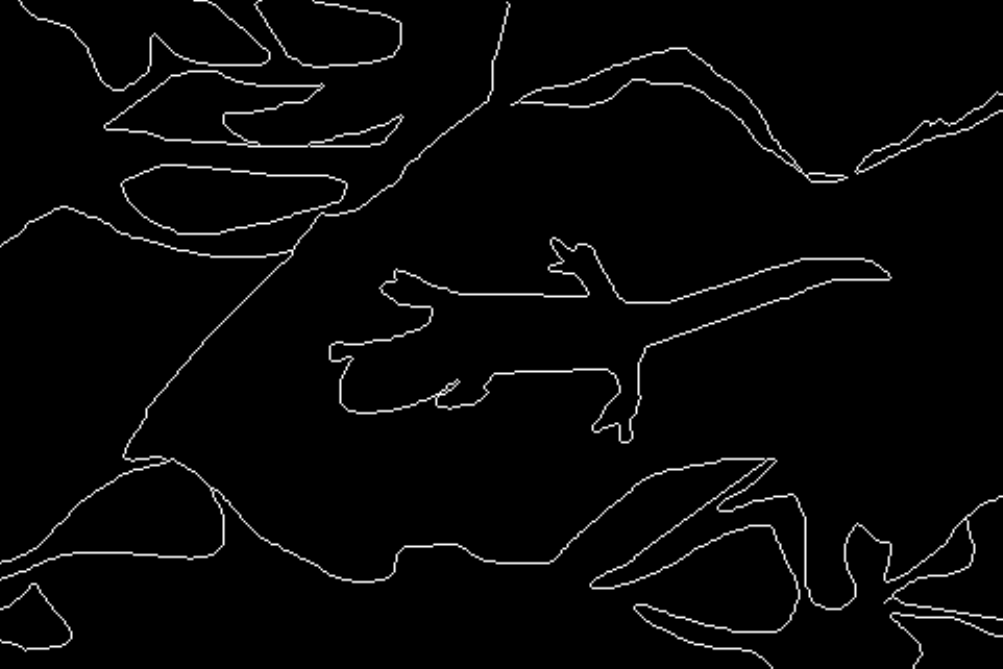} \label{fig:175083_b1} \\
& Ground-Truth II & \\
\includegraphics[width=0.23\paperwidth]		{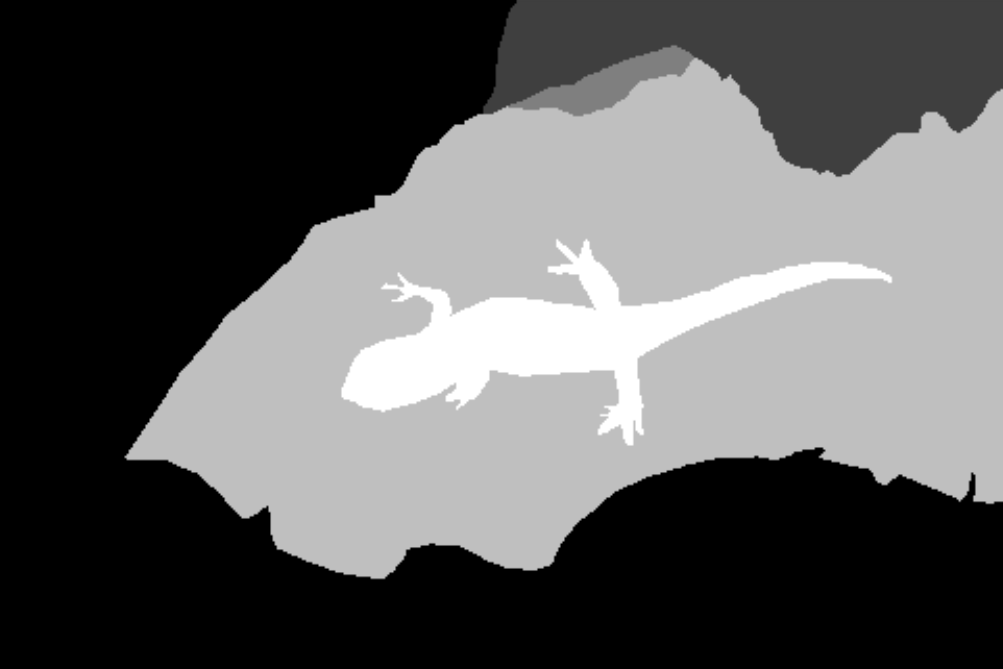} \label{fig:175083_2} &
\includegraphics[width=0.23\paperwidth]		{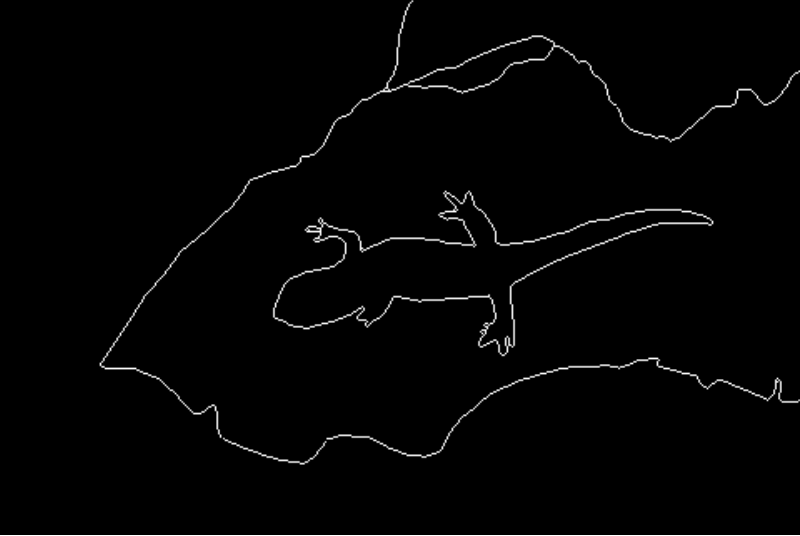} \label{fig:175083_a2} &
\includegraphics[width=0.23\paperwidth]		{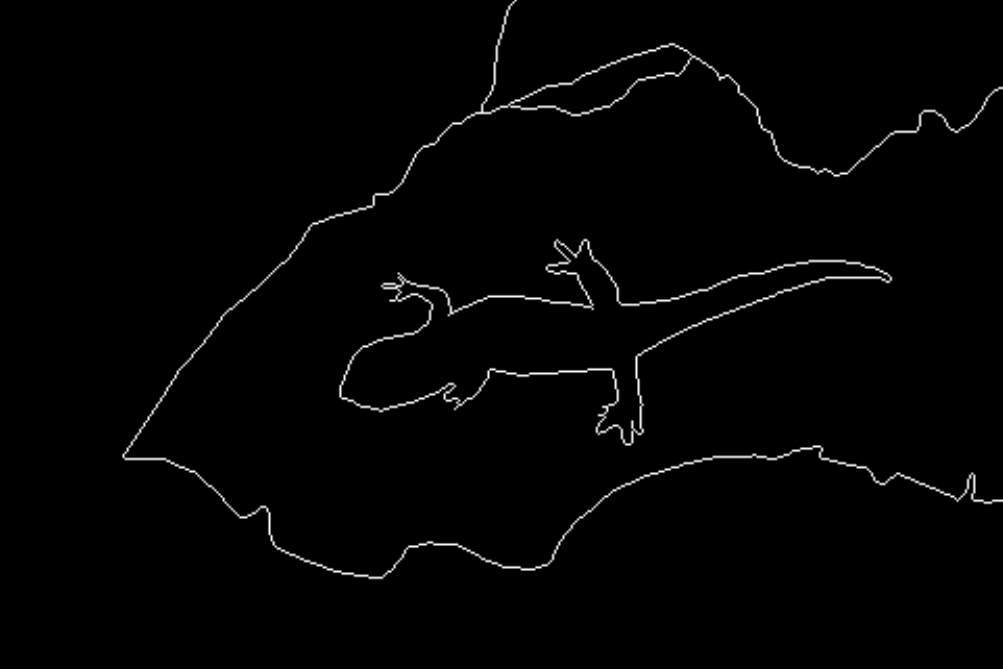} \label{fig:175083_b2} \\
& Ground-Truth III & \\
\includegraphics[width=0.23\paperwidth]		{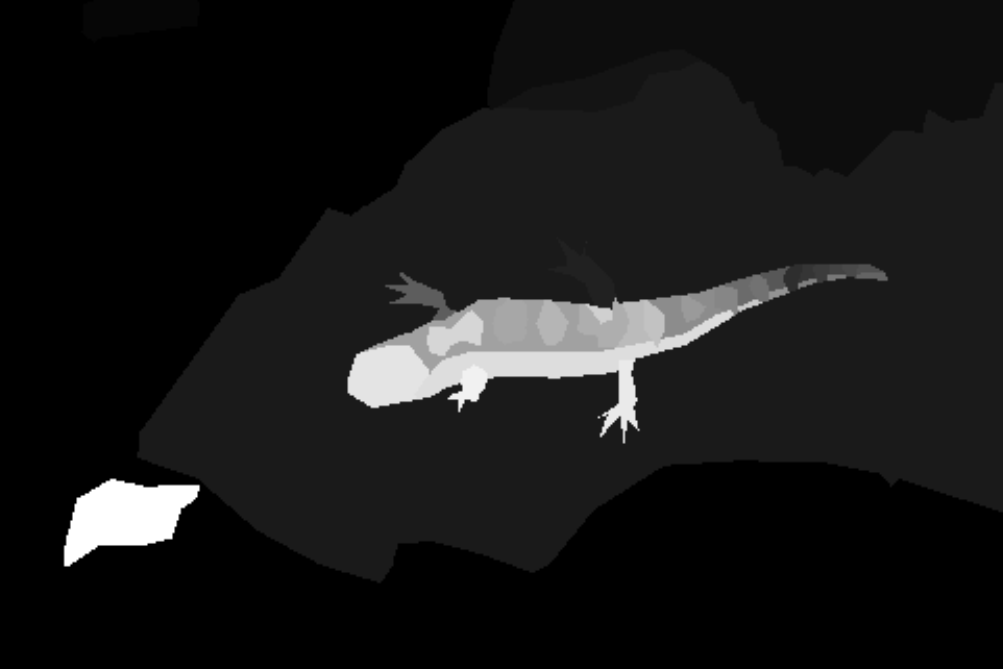} \label{fig:175083_3} &
\includegraphics[width=0.23\paperwidth]		{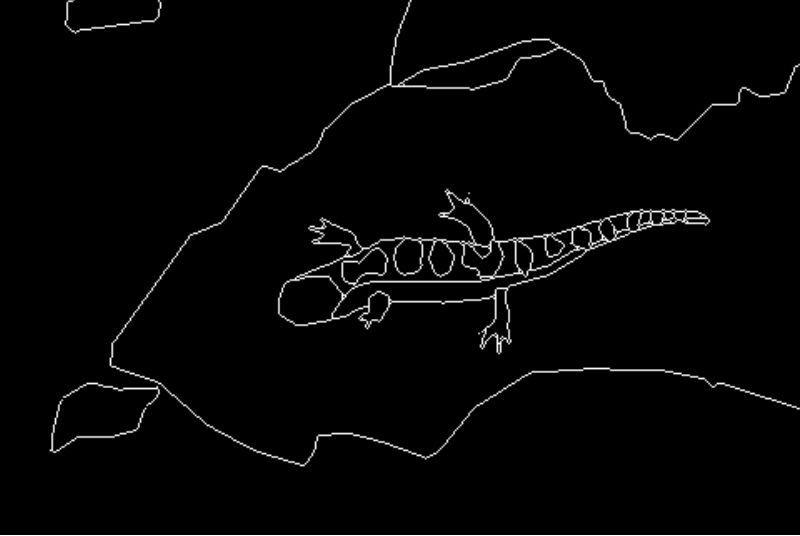} \label{fig:175083_a3} &
\includegraphics[width=0.23\paperwidth]		{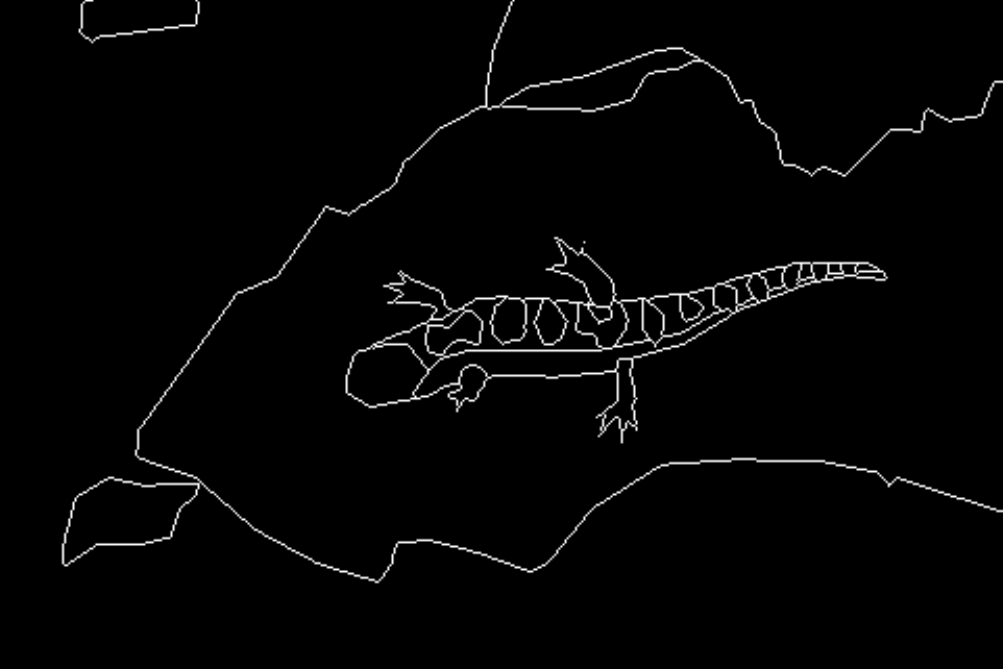} \label{fig:175083_b3} \\
& Ground-Truth IV & \\
\includegraphics[width=0.23\paperwidth]		{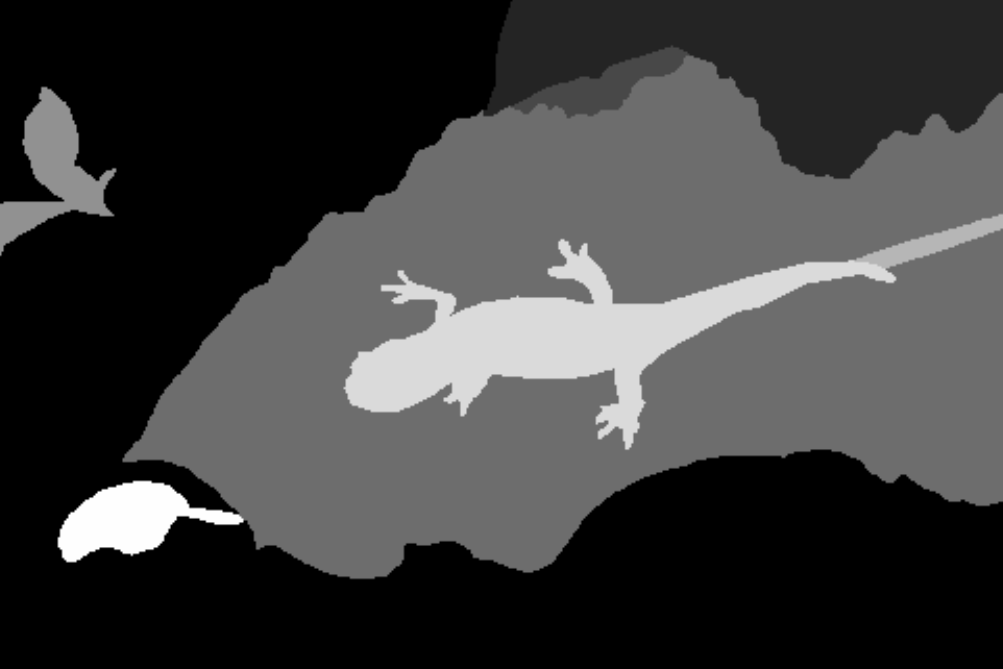} \label{fig:175083_5} &
\includegraphics[width=0.23\paperwidth]		{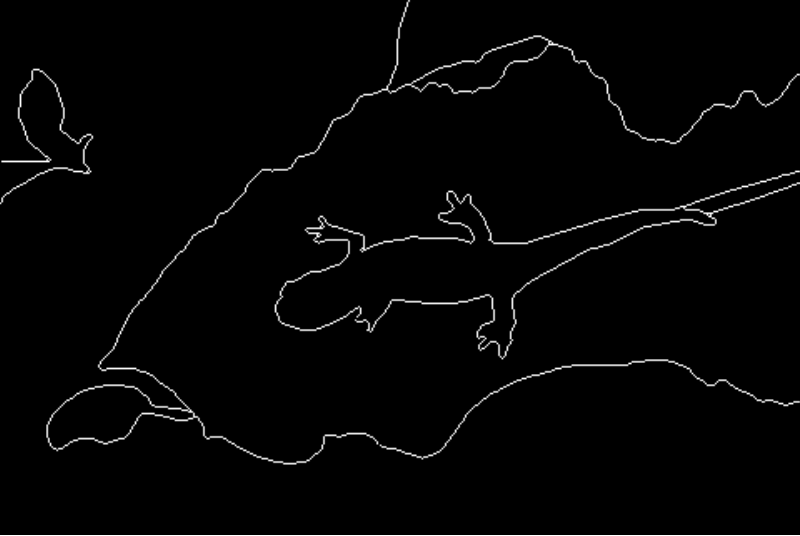} \label{fig:175083_a5} &
\includegraphics[width=0.23\paperwidth]		{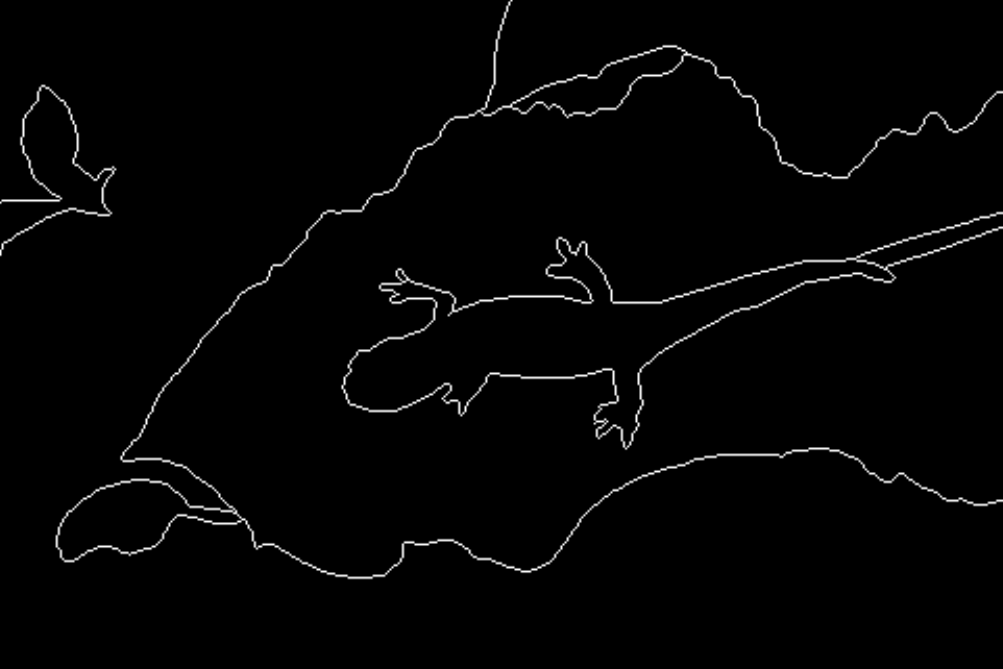} \label{fig:175083_b5} \\
& Ground-Truth V & \\
\includegraphics[width=0.23\paperwidth]		{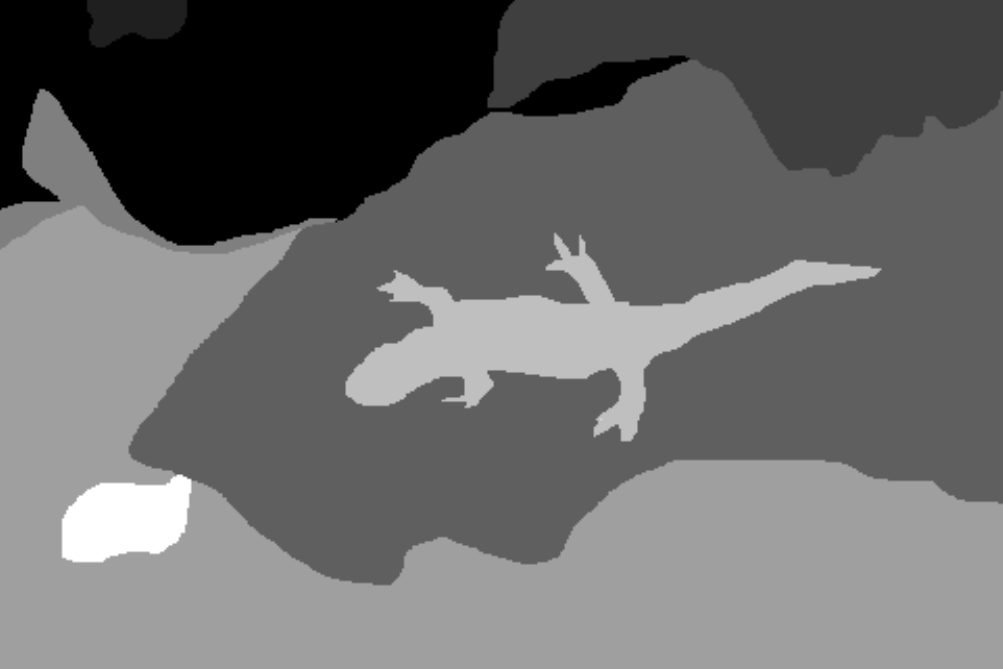} \label{fig:175083_6} &
\includegraphics[width=0.23\paperwidth]		{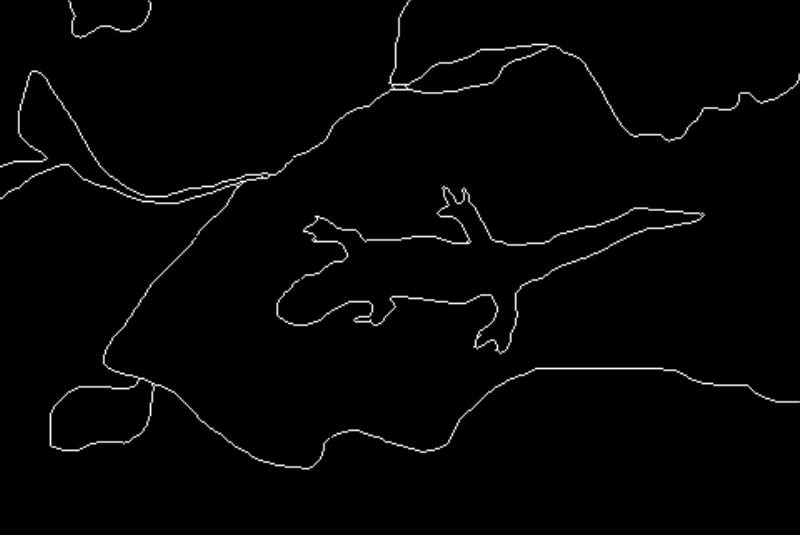} \label{fig:175083_a6} &
\includegraphics[width=0.23\paperwidth]		{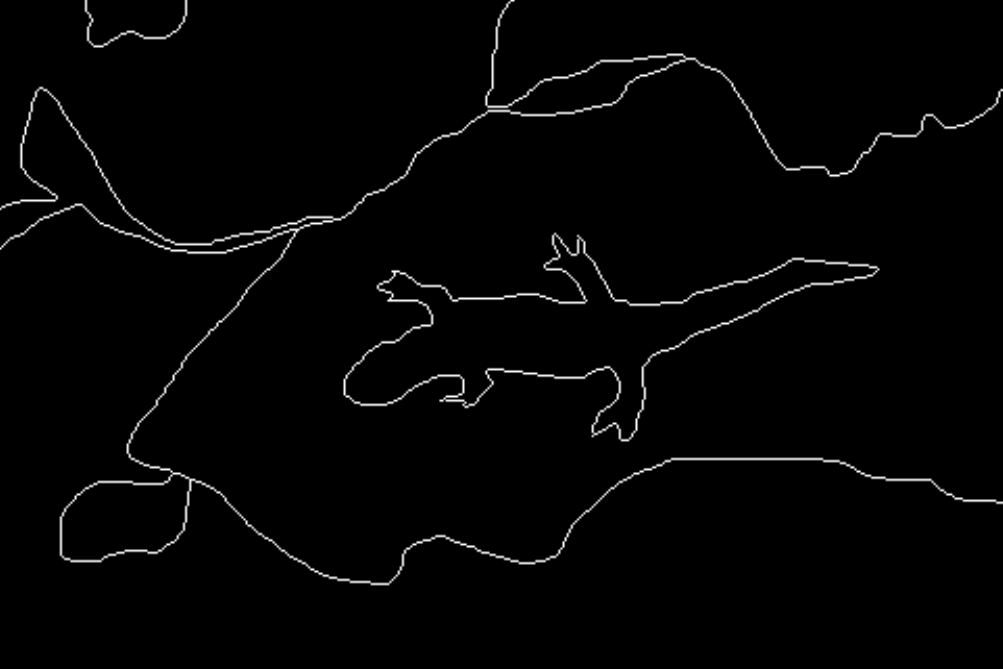} \label{fig:175083_b6}\\
\end{tabular}
\caption{Comparation of edge ground-truth segmentation with density algorithm segmentation. 
$1^{st}$ Column: Region ground-truth segmentation. $2^{nd}$ Column: Edge ground-truth segmentation. $3^{th}$ Column: Edge Detection by Density (EDD).}
\label{fig:experimentIII}
\end{figure}

\begin{table}[ht]
\centering
\vspace{-.05in}
\caption{Index of similarity among our edge detection algorithm and ground-truth images.}
\label{tab:experiment_results}
\begin{tabular}{c p{0.05cm} ccc p{0.05cm} ccc p{0.05cm} ccc}\hline
& & \multicolumn{3}{c}{Experiment I: Image 6303} & & \multicolumn{3}{c}{Experiment II: Image 41006} & & \multicolumn{3}{c}{Experiment III: Image 175083} \\
Ground-Truth	 & & PRI & NPRI & NED & & PRI & NPRI & NED & & PRI & NPRI & NED\\\hline
I. & & 0.968898 & 0.871664 & 0.087767 & & 0.960236 & 0.835922 & 0.128089 & & 0.928050 & 0.703113 & 0.190817\\
II. & & 0.971290 & 0.881534 & 0.088066 & & 0.956514 & 0.820564 & 0.122966 & & 0.949825 & 0.792963 & 0.144619\\
III. & & 0.969495 & 0.874127 & 0.061267 & & 0.934549 & 0.729930 & 0.171914 & & 0.939635 & 0.750916 & 0.138976\\
IV. & & 0.940551 & 0.754696 & 0.146649 & & 0.955903 & 0.818043 & 0.118532 & & 0.943945 & 0.768700 & 0.179005\\
V. & & 0.967766 & 0.866993 & 0.080747 & & 0.938192 & 0.744962 & 0.189659 & & 0.941919 & 0.760341 & 0.183362\\\hline
\multicolumn{13}{l}{Abbreviations: PRI: \textit{Probabilistic Rank Index}, NPRI: \textit{Normalized Probabilistic Rank Index}, NED: \textit{Natural Entropy Distance.}}
\end{tabular}
\vspace{-.1in}
\end{table}

\begin{figure}[ht]
\centering
\begin{tabular}{ccc}
Image 6303 of Berkeley's Database & Image 41006 of Berkeley's Database & Image 175083 of Berkeley's Database\\
\includegraphics[width=0.23\paperwidth, height=0.15\paperwidth]{images/segmentations/3063/3063_4}&
\includegraphics[width=0.23\paperwidth, height=0.15\paperwidth]{images/segmentations/41006/41006_3}&
\includegraphics[width=0.23\paperwidth, height=0.15\paperwidth]{images/segmentations/175083/175083_3} \\
& EDD & \\
\includegraphics[width=0.23\paperwidth, height=0.15\paperwidth]{images/segmentations/3063/3063_a4}& 
\includegraphics[width=0.23\paperwidth, height=0.15\paperwidth]{images/segmentations/41006/41006_a3}& 
\includegraphics[width=0.23\paperwidth, height=0.15\paperwidth]{images/segmentations/175083/175083_a3} \\
& Canny & \\
\includegraphics[width=0.23\paperwidth, height=0.15\paperwidth]{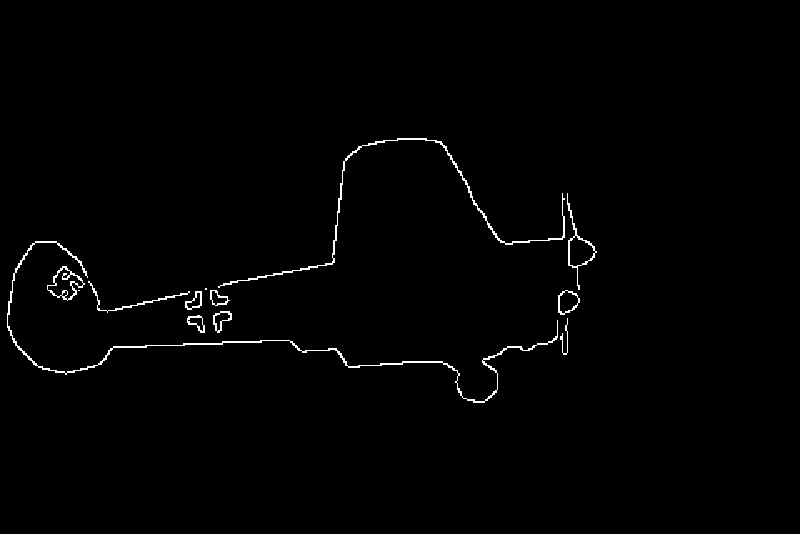}& 
\includegraphics[width=0.23\paperwidth, height=0.15\paperwidth]{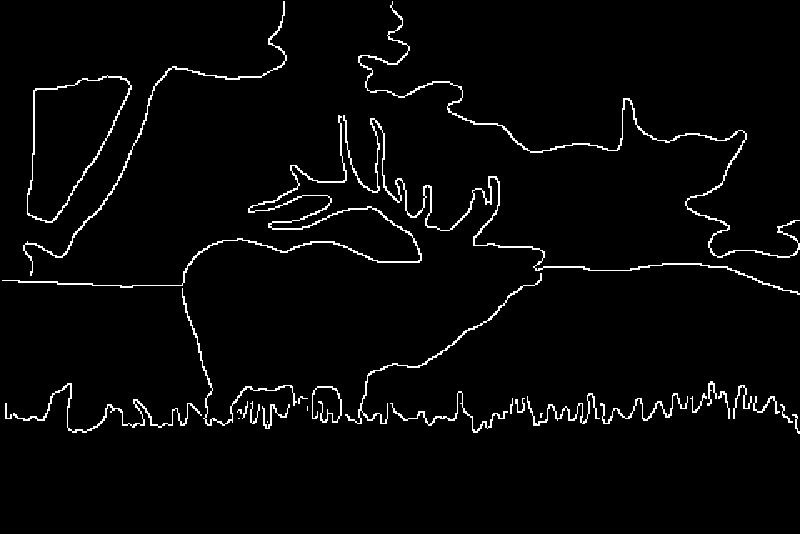}&
\includegraphics[width=0.23\paperwidth, height=0.15\paperwidth]{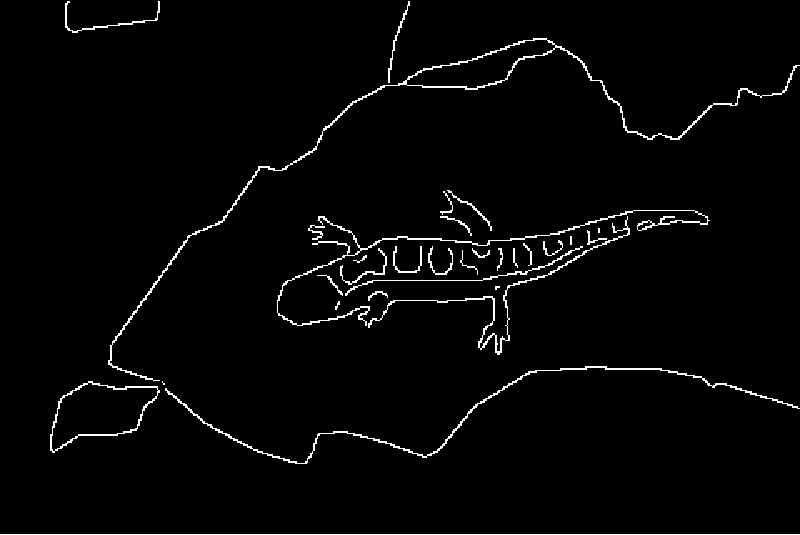}  \\
& Prewwit & \\
\includegraphics[width=0.23\paperwidth, height=0.15\paperwidth]{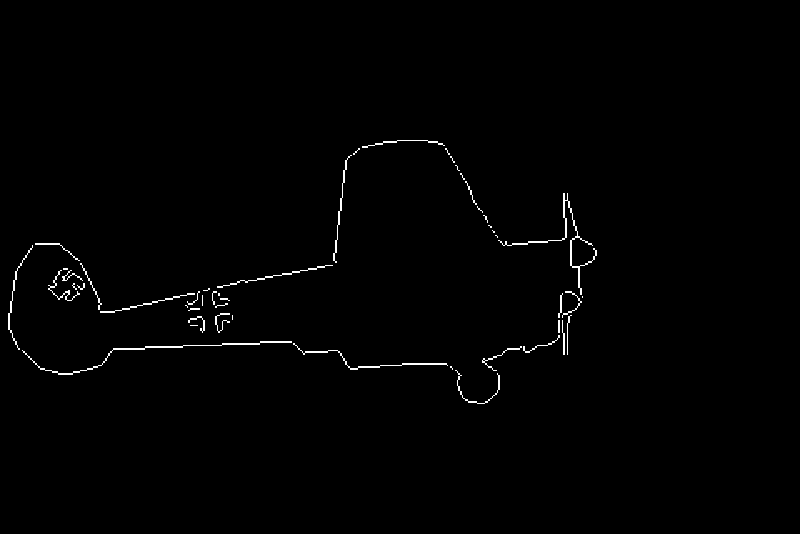}&
\includegraphics[width=0.23\paperwidth, height=0.15\paperwidth]{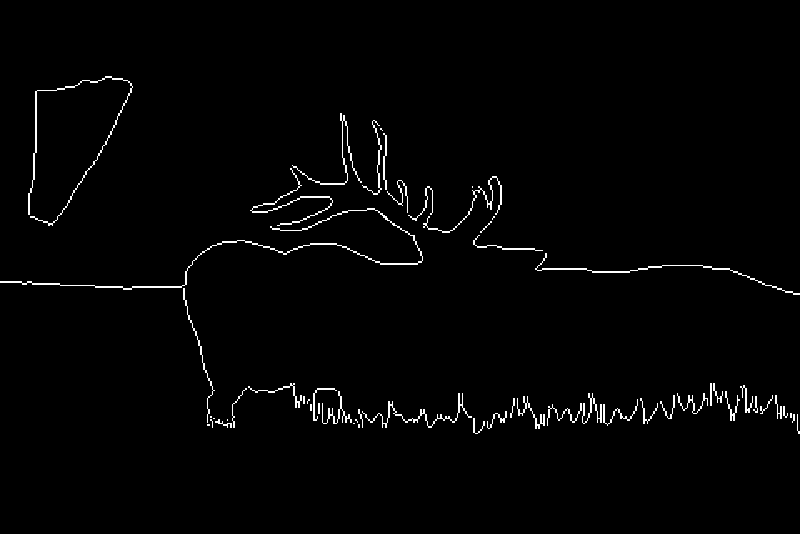}&
\includegraphics[width=0.23\paperwidth, height=0.15\paperwidth]{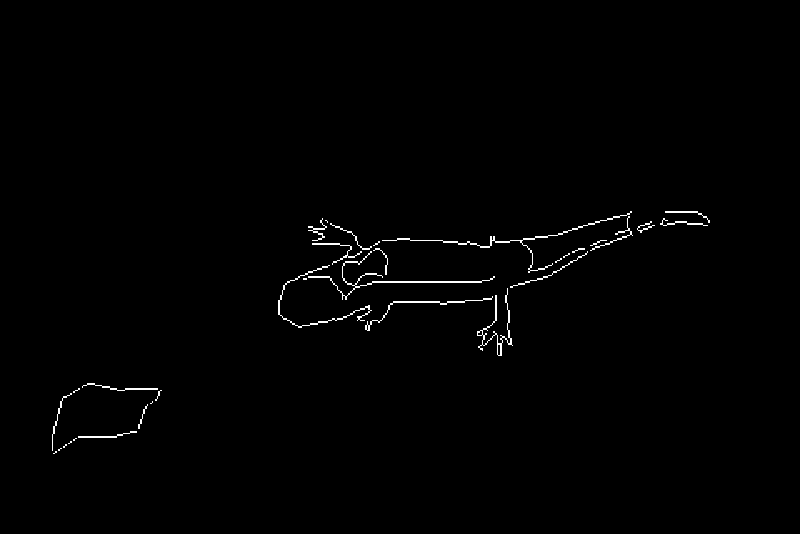} \\
& Roberts & \\
\includegraphics[width=0.23\paperwidth, height=0.15\paperwidth]{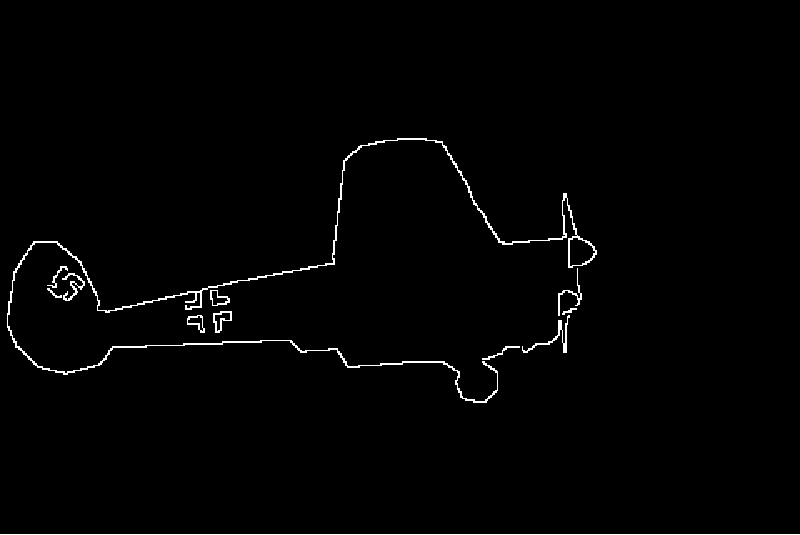}&
\includegraphics[width=0.23\paperwidth, height=0.15\paperwidth]{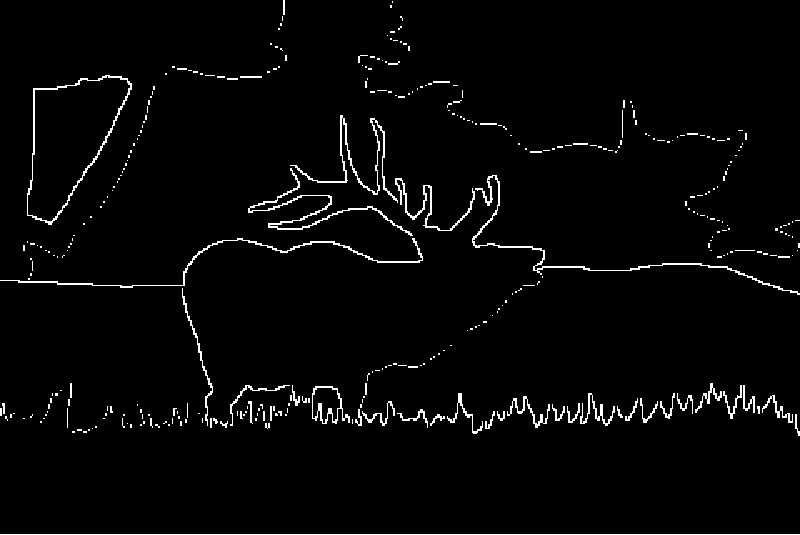}&
\includegraphics[width=0.23\paperwidth, height=0.15\paperwidth]{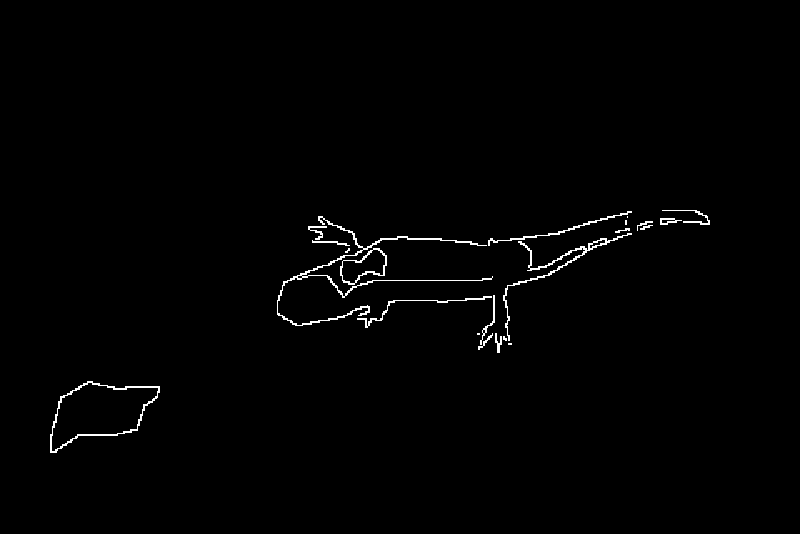} \\
& Sobel & \\
\includegraphics[width=0.23\paperwidth, height=0.15\paperwidth]{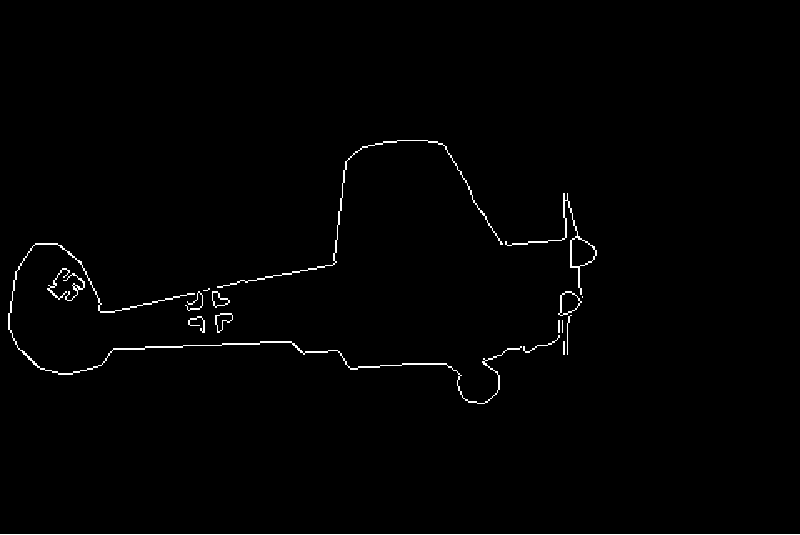}&
\includegraphics[width=0.23\paperwidth, height=0.15\paperwidth]{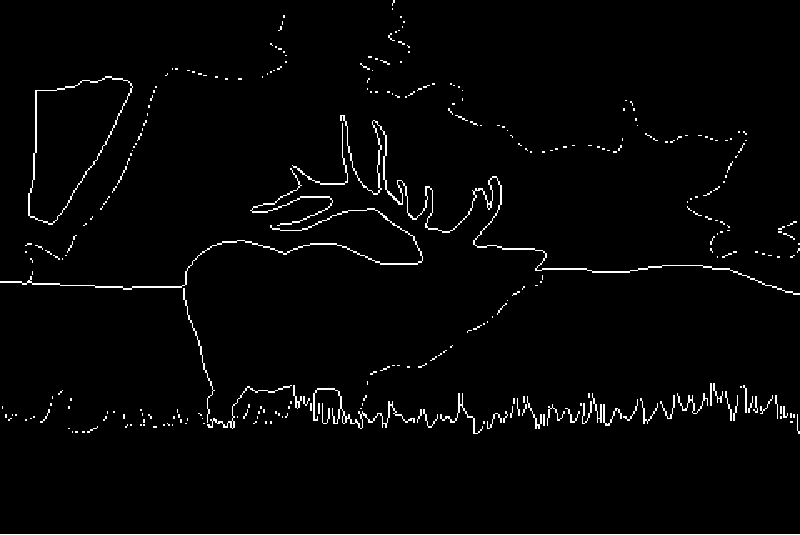}&
\includegraphics[width=0.23\paperwidth, height=0.15\paperwidth]{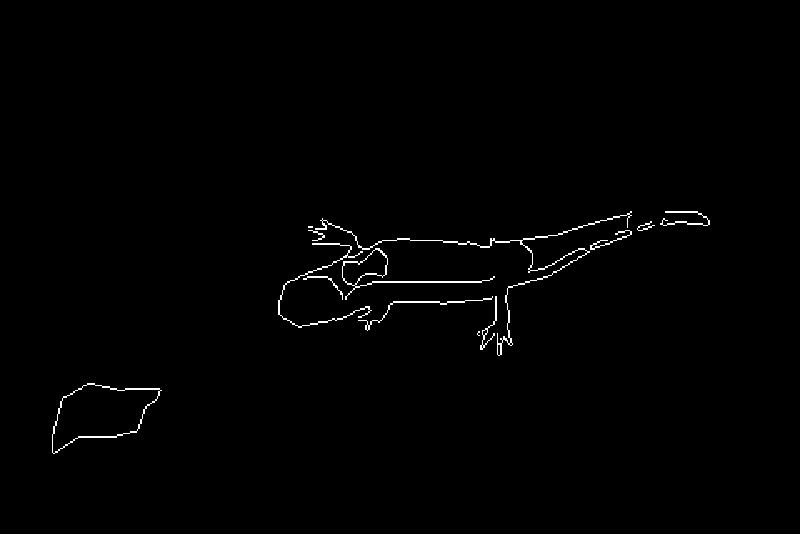}
\end{tabular}
\caption{Visual comparation of EDD with classical edge detectors.}
\label{fig:Comparation}
\end{figure}

\begin{table}[ht]
\vspace{-.1in}
\centering
\vspace{-.05in}
\caption{Quantitative comparation of EDD methods versus classical edge detector algorithms.}
\label{tab:experiment_comparation}
\begin{tabular}{c p{0.05cm} ccc p{0.05cm} ccc p{0.05cm} ccc}\hline
& & \multicolumn{3}{c}{Experiment I: Image 6303} & & \multicolumn{3}{c}{Experiment II: Image 41006} & & \multicolumn{3}{c}{Experiment III: Image 175083} \\
Algorithm & & PRI & NPRI & NED & & PRI & NPRI & NED & & PRI & NPRI & NED\\\hline
EDD & & 0.969495 & 0.874127 & 0.061267 & & 0.934549 & 0.729930 & 0.171914 & & 0.939635 & 0.750916 & 0.138976\\
Canny & & 0.976007 & 0.900997 & 0.035261 & & 0.941772 & 0.759734 & 0.101104 & & 0.945158 & 0.773706 & 0.080776\\
Prewitt & & 0.967537 & 0.866048 & 0.116485 & & 0.944864 & 0.772492 & 0.234304 & & 0.953138 & 0.806633 & 0.179450\\
Roberts & & 0.972056 & 0.884695 & 0.029136 & & 0.946424 & 0.778929 & 0.088014 & & 0.953728 & 0.809068 & 0.111554\\
Sobel & & 0.968576 & 0.870335 & 0.113957 & & 0.938396 & 0.745804 & 0.242177 & & 0.952704 & 0.804843 & 0.179254\\\hline
\multicolumn{13}{l}{Abbreviations: PRI: \textit{Probabilistic Rank Index}, NPRI: \textit{Normalized Probabilistic Rank Index}, NED: \textit{Natural Entropy Distance.}}\\
\end{tabular}
\end{table}

\begin{figure}[ht]
\centering
\subfigure[CT Image]{\includegraphics[width=0.26\paperwidth]{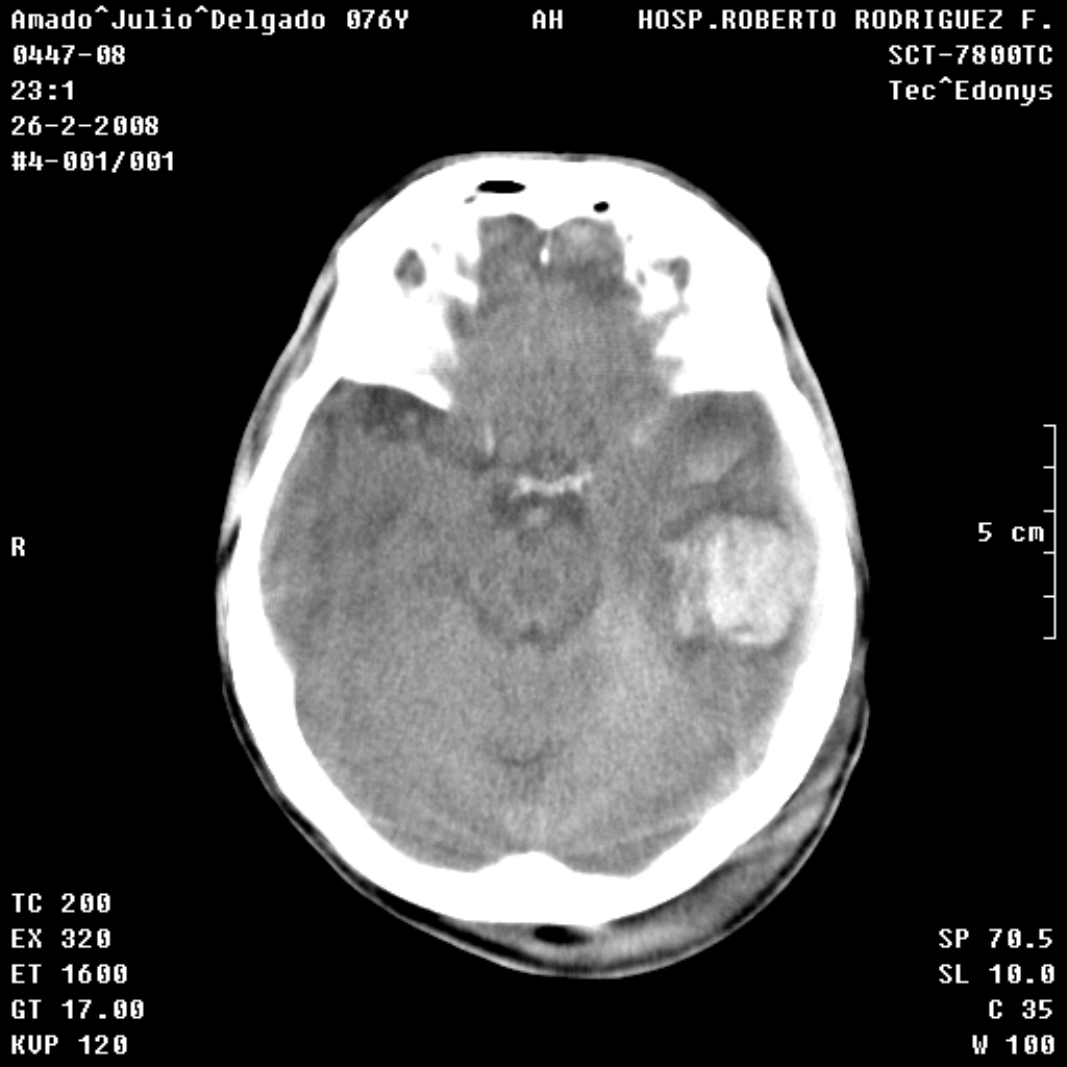}} 
\subfigure[\textit{MSHi} Image]{\includegraphics[width=0.26\paperwidth]{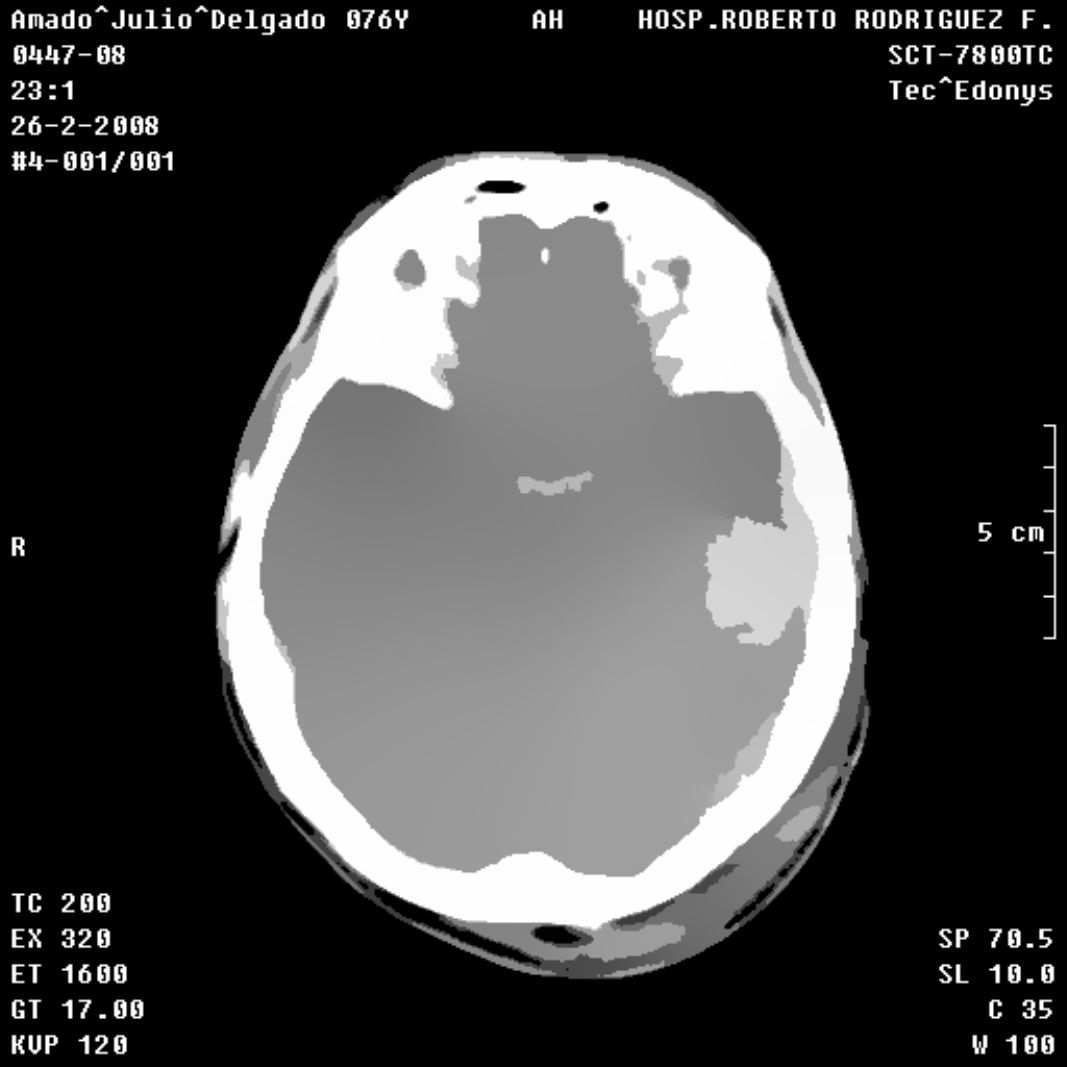}}  
\subfigure[hs=1 hr=15]{\includegraphics[width=0.26\paperwidth]{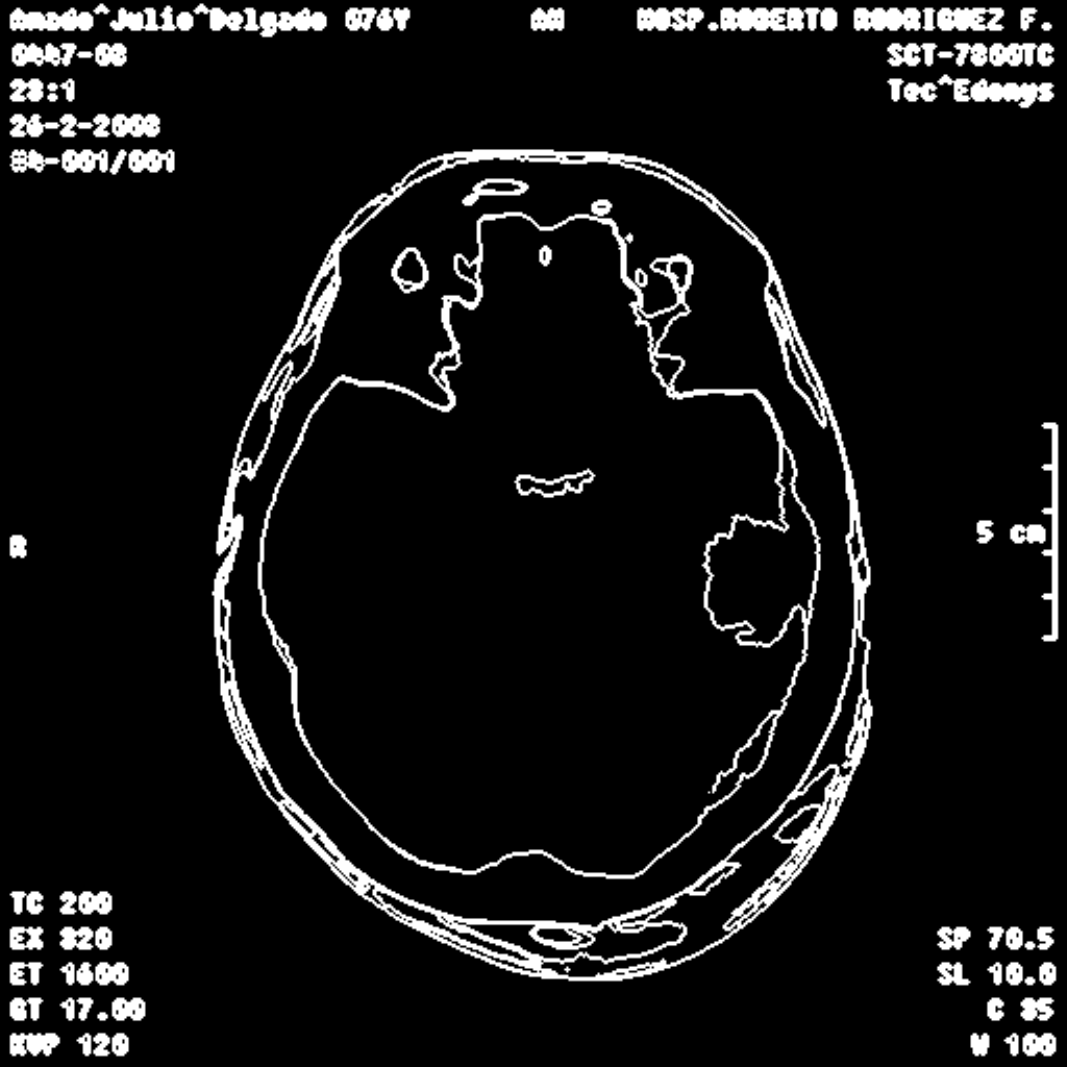}} \\
\subfigure[hs=3 hr=15]{\includegraphics[width=0.26\paperwidth]{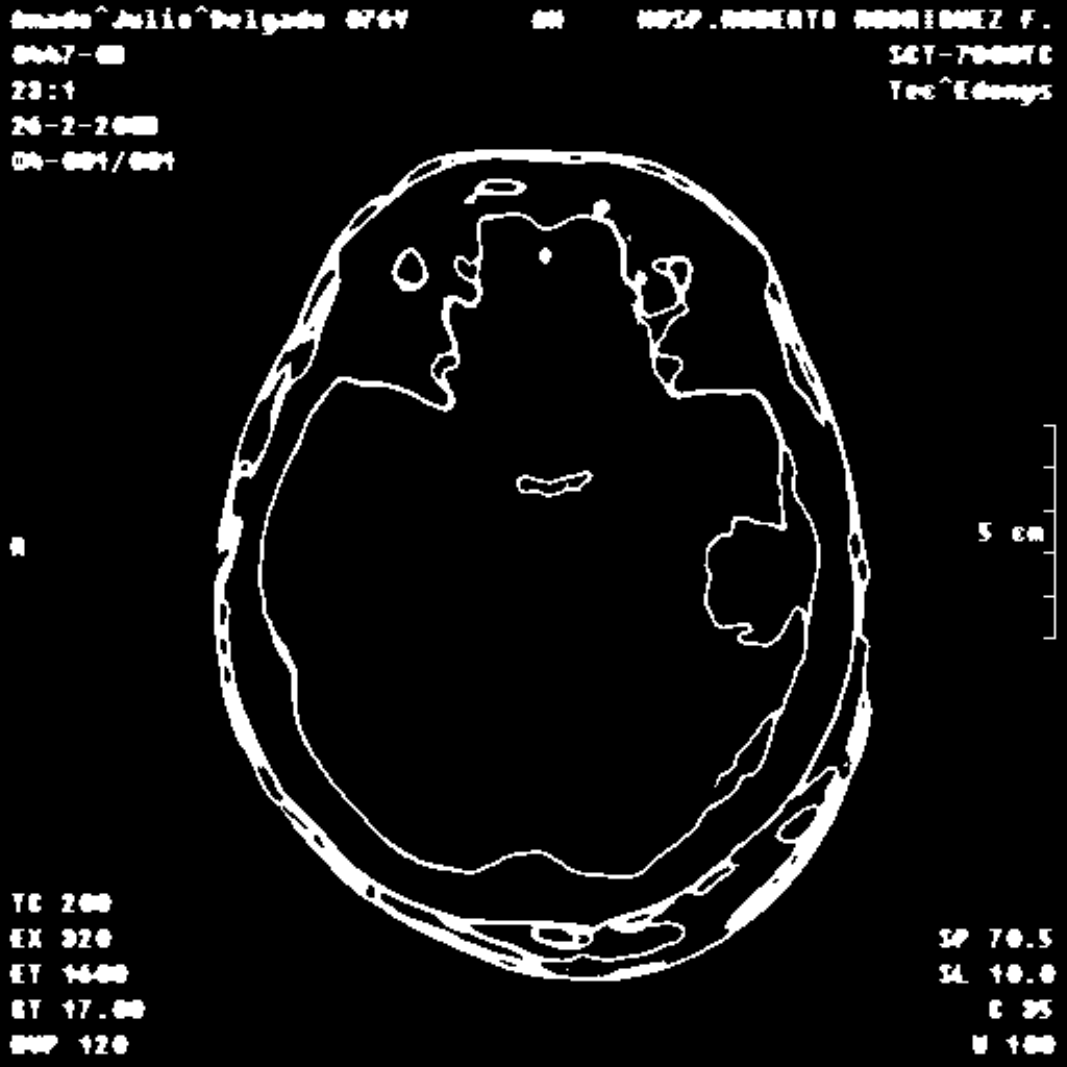}} 
\subfigure[hs=5 hr=15]{\includegraphics[width=0.26\paperwidth]{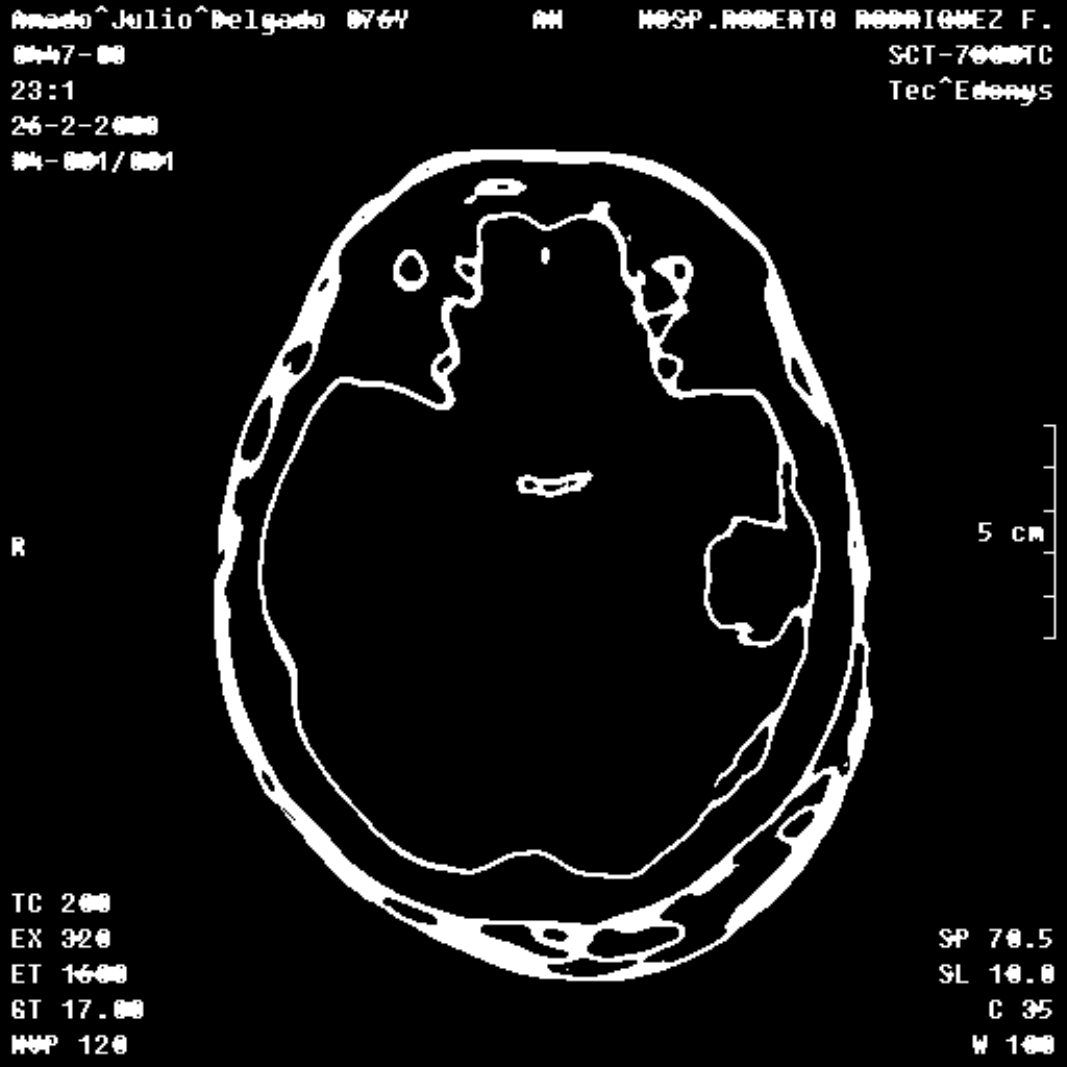}} 
\subfigure[hs=7 hr=15]{\includegraphics[width=0.26\paperwidth]{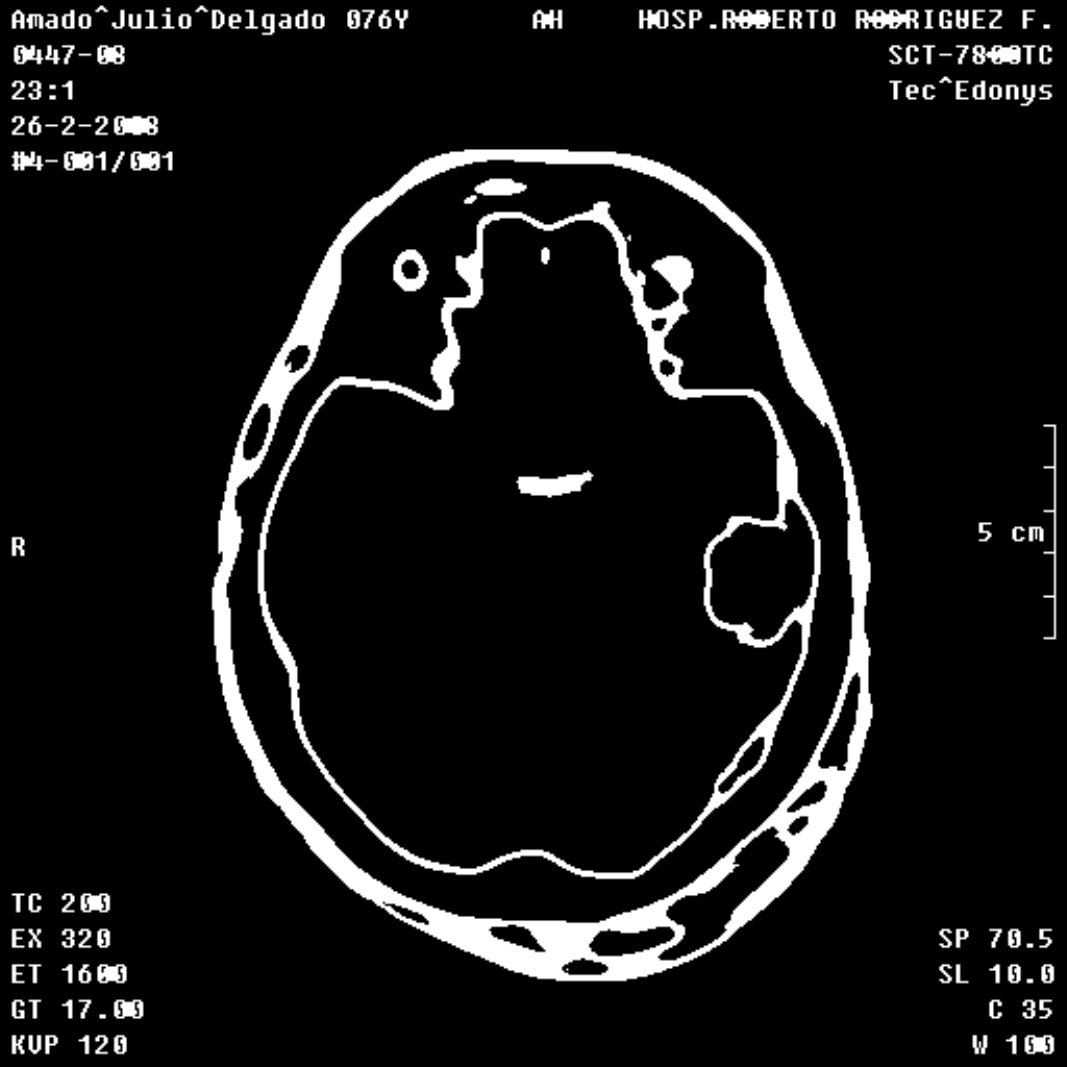}} \\
\subfigure[hs=1 hr=5]{\includegraphics[width=0.26\paperwidth]{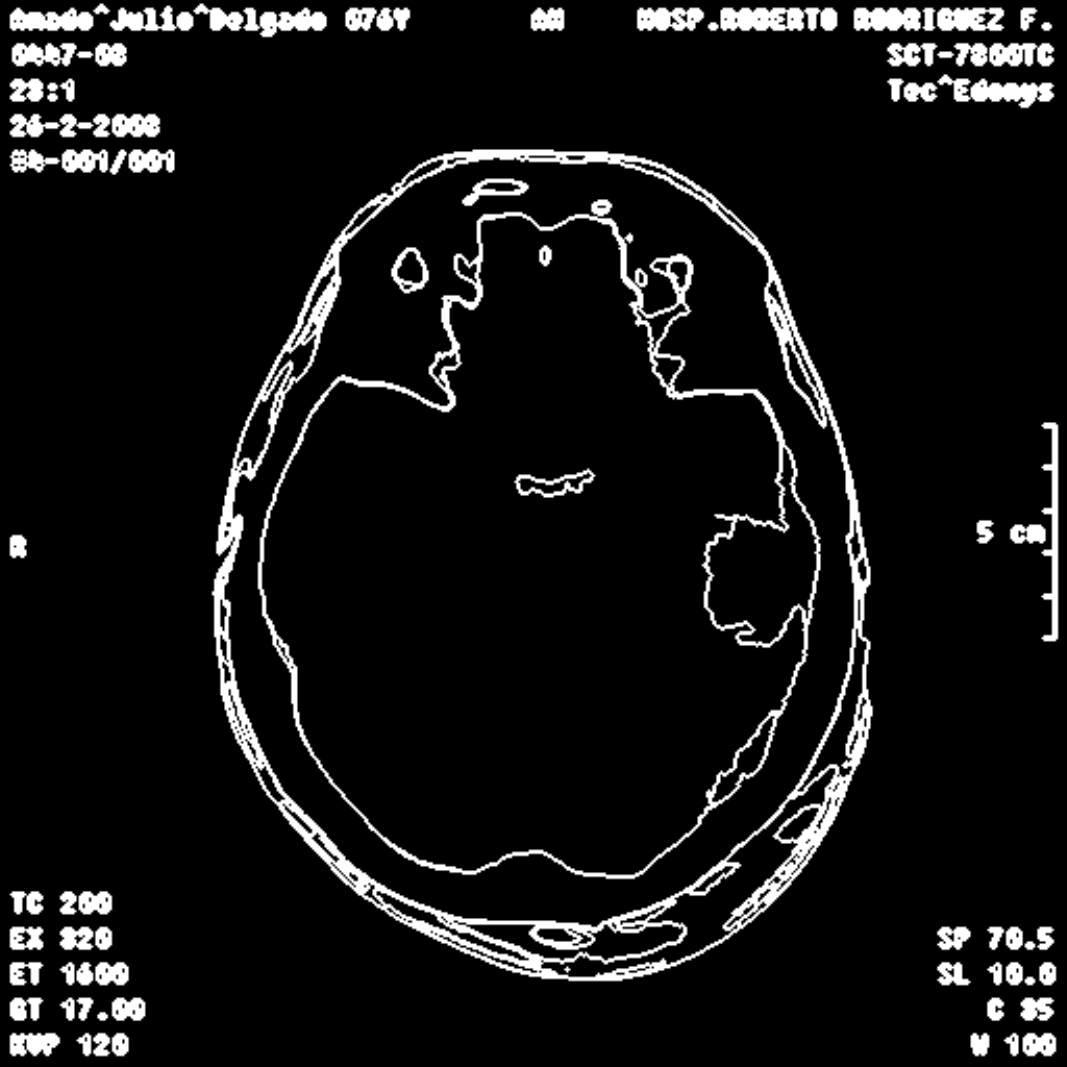}} 
\subfigure[hs=1 hr=7]{\includegraphics[width=0.26\paperwidth]{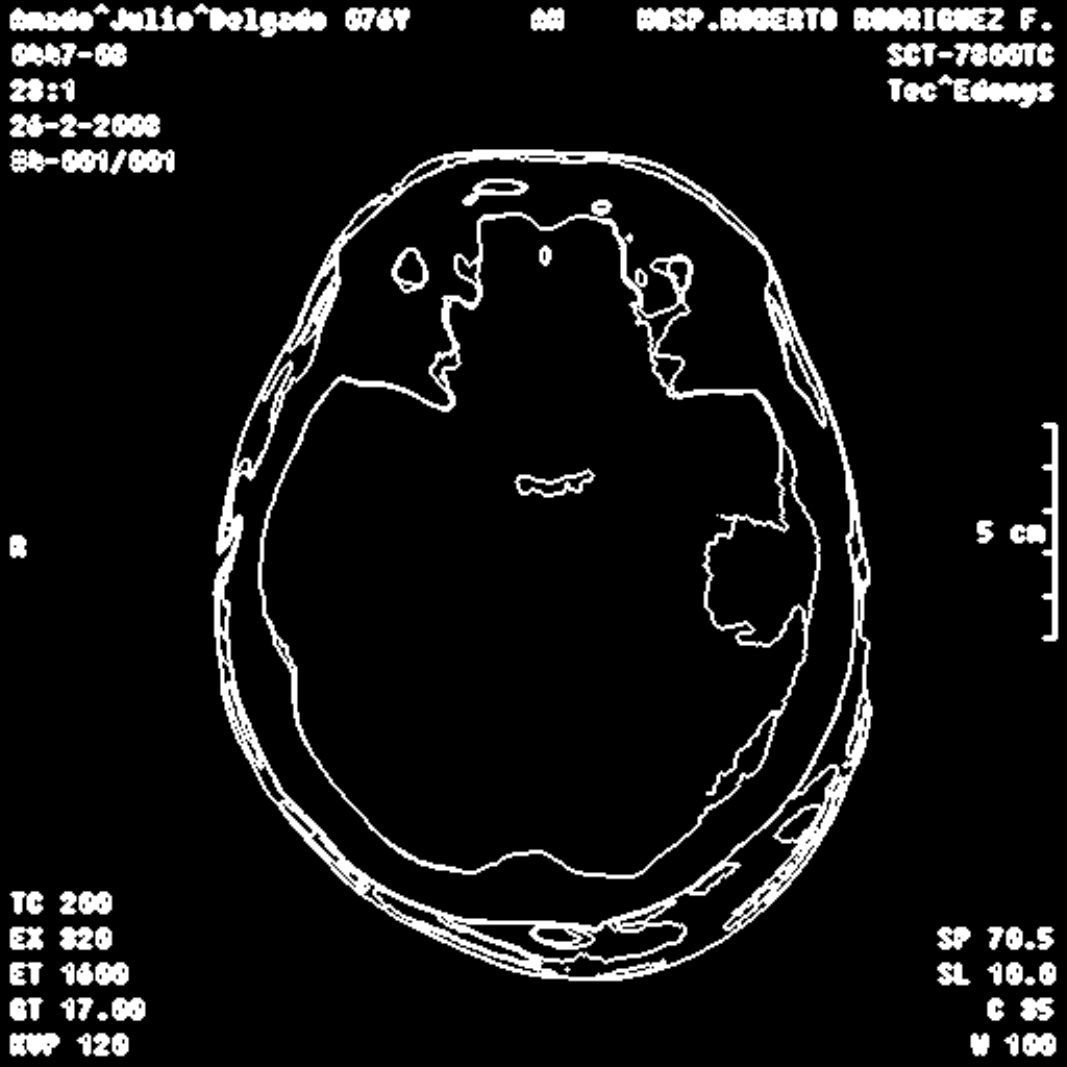}} 
\subfigure[hs=1 hr=30]{\includegraphics[width=0.26\paperwidth]{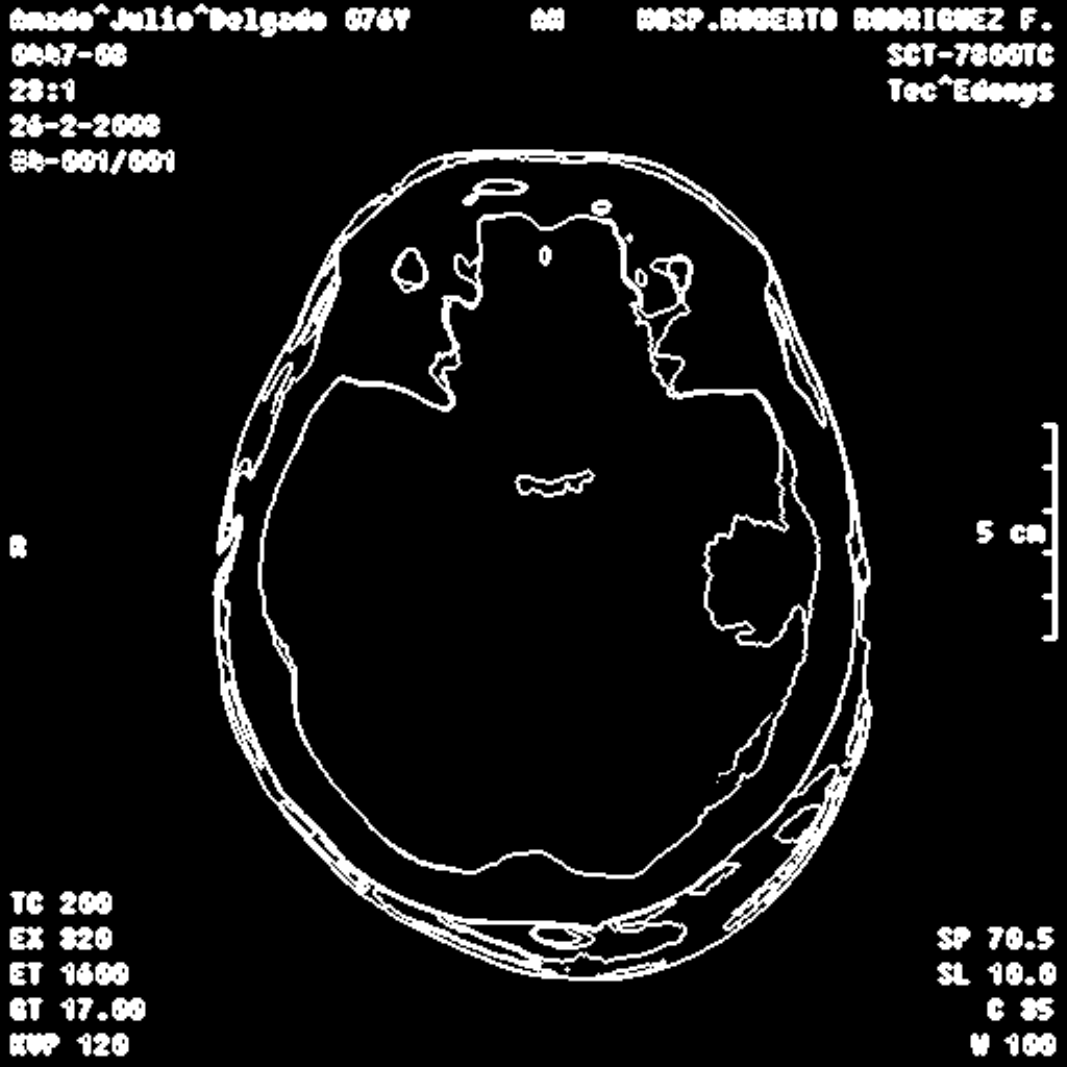}} 
\caption{Examples of Density Edge Detection in a CT image segmented with \textit{MSHi} algorithm.}
\label{fig:Results}
\end{figure}

\end{document}